\newif\ifready\readytrue
\newif\ifarxiv\arxivtrue

\ifready
\else

\fi

\documentclass[10pt]{article}

\ifarxiv

\usepackage{fullpage}
\usepackage{times}
\usepackage{xurl}
\usepackage{natbib}

\usepackage{lscape}
\usepackage{rotating}
\usepackage{caption}
\usepackage{adjustbox}

\date{}

\else

\usepackage[final]{neurips_2024}
\usepackage{caption}
\usepackage{adjustbox}

\fi

\usepackage{amsthm,amsmath,amsfonts,amssymb}
\usepackage{felix}
\usepackage{tikz}
\usepackage{tikz-cd}
\usepackage{quiver}
\newcommand{\OWS}{\ensuremath{\mathcal{OWS}}\xspace}
\newcommand{\ComputeForward}{\ensuremath{\mathtt{ComputeForward}}\xspace}
\newcommand{\rRound}{\ensuremath{\mathtt{rRound}}\xspace}
\newcommand{\rQuantileEst}{\ensuremath{\mathtt{rQuantileEst}}\xspace}
\newcommand{\rLearnerOWSd}{\ensuremath{\mathtt{rLearnerOWSd}}\xspace}

\newcommand{\rInflEst}{\ensuremath{\mathtt{rInflEst}}\xspace}
\newcommand{\rBuildDT}{\ensuremath{\mathtt{rBuildDT}}\xspace}
\newcommand{\rLift}{\ensuremath{\mathtt{rLift}}\xspace}

\newcommand{\rFiniteDistrEst}{\ensuremath{\mathtt{rFiniteDistrEst}}\xspace}
\newcommand{\rBoost}{\ensuremath{\mathtt{rBoost}}\xspace}

\DeclareMathOperator{\Infl}{Infl}

\author{%
 \begin{tabular}{cc}	
        \begin{tabular}{c}
        \text{Alkis Kalavasis} \\
		 Yale University \\
		\href{mailto:alvertos.kalavasis@yale.edu}{alvertos.kalavasis@yale.edu}
        \end{tabular}
        & 	
		\begin{tabular}{c}
		\text{Amin Karbasi} \\
		 Yale University \\
		\href{mailto:amin.karbasi@yale.edu}{amin.karbasi@yale.edu}
		\end{tabular}
		\\
		\\
		\begin{tabular}{c}
		\text{Grigoris Velegkas} \\
		Yale University \\
		\href{mailto:grigoris.velegkas@yale.edu}{grigoris.velegkas@yale.edu}
		\end{tabular}
		 & 
		 \begin{tabular}{c}
		 \text{Felix Zhou} \\
		 Yale University \\
		\href{mailto:felix.zhou@yale.edu}{felix.zhou@yale.edu}\\
		\end{tabular}
		\end{tabular}
}

\title{On the Computational Landscape of \\ Replicable Learning}

\begin{document}

\maketitle

\begin{abstract}
    \sloppy
    We study
    computational aspects of algorithmic replicability, 
    a notion of stability introduced
    by \citet*{impagliazzo2022reproducibility}. 
    Motivated by a recent
    line of work that established strong \emph{statistical} connections between replicability and other notions
    of learnability such as online learning, private learning, and SQ learning, we aim to understand better the \emph{computational}
    connections between replicability and these learning paradigms.
    Our first result shows that there is a concept class that is efficiently replicably PAC learnable,
    but, under standard cryptographic assumptions, no efficient
    online learner exists for this class. 
    Subsequently, we design an efficient replicable
    learner for PAC learning parities when the marginal
    distribution is far from uniform, making progress on a question posed by \citet{impagliazzo2022reproducibility}.
    To obtain this result, we design a replicable lifting framework
    inspired by \citet*{blanc2023lifting}
    that transforms in a black-box manner efficient replicable PAC learners under the uniform marginal distribution over the Boolean hypercube
    to replicable PAC learners under any marginal distribution,
    with sample and time complexity that depends on a certain measure of the complexity of the distribution.
    Finally,
    we show that any pure DP learner can be transformed 
    to a replicable one in time polynomial in the accuracy, confidence parameters and exponential in the representation dimension of the underlying hypothesis class.
\end{abstract}

\section{Introduction}
The replicability crisis is omnipresent in many scientific disciplines including
biology, chemistry, and, importantly, AI \citep{baker20161,pineau2019iclr}. A recent article
that appeared in Nature \citep{ball2023ai}
explains how the reproducibility crisis witnessed in AI
has a cascading effect across many other scientific areas
due to its widespread applications in other fields, like medicine.
Thus, 
a pressing task is to design
a formal framework through which we can argue about the replicability of experiments in ML. 
Such an attempt was initiated recently
by the pioneering work of \citet{impagliazzo2022reproducibility},
who proposed a definition of replicability as a property
of learning algorithms.
\begin{restatable}
[Replicable Algorithm; \citealp{impagliazzo2022reproducibility}]
{defn}
{replicableDefinition}
\label{def:replicable}
    Let $\mathcal R$ be a distribution over random strings. A learning algorithm
    $\mathcal A$ is $n$-sample $\rho$-replicable under distribution $\mathcal D$ if for two independent sets $S, S' \sim \mathcal D^n$ it holds
    that $\Pr_{S,S' \sim \mathcal D^n, r \sim \mathcal R} [\mathcal A(S,r) \neq \mathcal A(S',r)] \leq \rho$. We will say that $\mathcal A$ is replicable if the above holds uniformly over all distributions $\mathcal D$.
\end{restatable}
We emphasize that the random string $r$ is shared across the two executions and this aspect of the definition is crucial
in designing algorithms that have the same output
under two different i.i.d. inputs.
Indeed, both \citet{dixon2023list} and \citet{chase2023replicability} demonstrated learning tasks
for which there are no algorithms
that satisfy this strong notion of replicability 
when the randomness is not shared across executions. 
This shared random string models the random seed of a learning algorithm in practice,
and sharing internal randomness can be easily implemented by sharing said random seed. 

Closer to our work, \citet{impagliazzo2022reproducibility,ghazi2021user,bun2023stability,kalavasis2023statistical} established strong \emph{statistical} connections between replicability and other notions of algorithmic stability and learning paradigms such as differential privacy (DP), statistical queries (SQ), and online learning.
{Although several of these works provided various \emph{computational} results, these connections are not as well understood as the statistical ones.}

\subsection{Our Contributions}
In this work, we aim to shed further light on the {aforementioned \emph{computational} connections}. In particular, we provide both negative and positive results. Negative results are manifested through \emph{computational separations} while positive results are presented through \emph{computational transformations}, as we will see shortly.
We emphasize that whenever we work within the PAC learning framework we consider the \emph{realizable} setting.

\paragraph{Replicability \& Online Learning.}
The results of \citet{ghazi2021user, bun2023stability, kalavasis2023statistical} established a statistical connection between replicability and online learning.
In particular, in the context of PAC learning, 
these works essentially show that replicable PAC learning is statistically equivalent to online learning
since learnability in both settings
is characterized by the finiteness of the Littlestone 
dimension of the underlying concept class.
From the above, the first natural question is the following:
\begin{center}
Q1. \emph{How does replicability computationally relate to online learning?}    
\end{center}
We show that under standard cryptographic assumptions, efficient replicability is separated from efficient online learning.
\begin{thm}
[Informal, see \Cref{thm:main1}]
    Assuming the existence of one-way functions, there
    is a concept class that is replicably PAC learnable in polynomial time, 
    but is not {efficiently} online learnable by
    any no-regret algorithm.
\end{thm}

In order to prove the above result,
we provide an efficient replicable PAC learner for the concept class of \emph{One-Way Sequences} \OWS over $\{0,1\}^d$, introduced by \citet{blum1994separating}. This algorithm, 
which relies on a novel replicable subroutine for quantile estimation (cf. \Cref{thm:replicable quantile est}),
combined with the cryptographic hardness result of \citet{bun2020computational} for online learnability,
gives the desired computational separation. For further details, we refer
to \Cref{sec:online}.

\paragraph{Replicability \& SQ.} The Statistical Query (SQ) framework is an expressive model of statistical learning introduced by \citet{kearns1998efficient}. 
This model of learning is a restricted class of algorithms that 
is only permitted indirect access to samples through approximations of certain carefully chosen functions of the data
(statistical queries).
The motivation behind the introduction of the SQ framework was to capture a large class of noise-resistant learning algorithms.
As such, any SQ algorithm enjoys some inherent stability properties. Since replicability also captures some notion of algorithmic stability, it is meaningful to ask:
\begin{center}
    Q2. \emph{How does replicability computationally relate to SQ learning?}
\end{center}
For some background about the SQ framework, we refer the reader to \Cref{sec:background}.
A manifestation of the intuitive connection between SQ and replicability can be found in 
one of the main results of \citet{impagliazzo2022reproducibility}, which states that any efficient SQ algorithm can be efficiently made replicable. In this paper,
we investigate the other direction of this connection.

A particularly interesting computational separation between (distribution-specific) replicability and SQ learning
was noticed by \citet{impagliazzo2022reproducibility}, in the context of realizable binary classification: if we consider the uniform distribution $\mcal U$ over the Boolean hypercube $\{0,1\}^d$, then
the concept class of parities  is SQ-hard to learn under $\mcal U$ \citep{kearns1998efficient} but admits an efficient PAC learner that is replicable under $\mcal U$.
Indeed, 
with high probability over the random draw of the data,
Gaussian elimination, 
which is the standard algorithm for PAC learning parities under $\mcal U$,
gives a \emph{unique} solution and, as a result, is replicable. 

Based on this observation, \citet{impagliazzo2022reproducibility}
posed as an interesting question whether parities are efficiently learnable by a replicable learner under other marginal distributions, for which Gaussian elimination is ``unstable'', i.e., it fails to give a unique solution (see e.g., \Cref{prop:Gaussian elimination fails}). 

\paragraph{Transforming Distribution-Specific Replicable Learners.} Inspired by the question of \citet{impagliazzo2022reproducibility}, 
we ask
the following more general question for realizable binary classification:
\begin{center}
    Q3. \emph{For a class $\mathscr C \subseteq \{0,1\}^\mcal X$ and marginal distributions $\mcal D_1, \mcal D_2$ over $\mcal X$,
    how does\\replicable PAC learnability of $\mathscr C$ under $\mathcal{D}_1$ and under $\mcal D_2$ relate to one another computationally?}
\end{center}

Our main result is a general black-box transformation from a replicable PAC learner under the uniform distribution over $\mcal X = \{0,1\}^d$ to a replicable PAC learner under some unknown marginal distribution $\mathcal{D}$. 
The runtime of the transformation depends on the \emph{decision tree complexity} of the distribution  $\mathcal D$, 
a complexity measure that comes from the recent work of \citet{blanc2023lifting} and is defined as follows:
\begin{defn}
[Decision Tree Complexity; \citealp{blanc2023lifting}]
\label{def:DTmain} 
    The {decision tree complexity} of a distribution $\mathcal{D}$ over $\{0,1\}^d$ is the smallest integer $\ell$ such that its probability mass function (pmf) can be computed by a depth-$\ell$ decision tree (cf. \Cref{defn:decision tree}). 
\end{defn} 
A \emph{decision tree} (cf. \Cref{defn:decision tree}) $T : \{0,1\}^d \to \R$ is a binary tree whose internal nodes query a particular coordinate $x_i, i \in [d]$ (descending left if $x_i = 0$ and right otherwise) and whose leaves are labelled by real values.
Each $x\in \Set{0, 1}^d$ follows a unique root-to-leaf path based on the queried coordinate of internal nodes,
and its value $T(x)$ is the value stored at the root.
For some $\mcal D$ over the Boolean hypercube, 
we say $\mcal D$ is \emph{computed by a decision tree $T$ of depth $\ell$} if, 
for any $x \in \{0,1\}^d$, 
it holds that $\mcal D(x) = T(x)$.

As an example,
note that the uniform distribution requires depth $\ell = 1$ since its pmf is constant while a distribution whose pmf takes $2^d$ different values requires depth $\ell = d$. 
However, 
various natural (structured) distributions have tree complexity much smaller than $d$. Importantly, the running time overhead of our lifting approach scales proportionally to the decision tree complexity of $\mcal D$ and hence can be used to obtain novel efficient replicable PAC learners. 

Our replicable lifting framework informally states the following:
\begin{thm}
[Informal, see \Cref{thm:lift}]
\label{thm:main4}
    Let $\mathcal X = \{0,1\}^d$ and $m = \poly(d)$.
    Consider a concept class $\mathscr C \subseteq \{0,1\}^{\mathcal X}$ and assume that $\mathcal A$ is a PAC learner for $\mathscr C$ that is $m$-sample replicable under the uniform distribution and runs in time $\poly(m)$.
    Then, for some (unknown) monotone\footnote{Recall that $\mcal D$ over $\{0,1\}^d$ is monotone if whenever $x \succeq y$ in the partial order of the poset, we have $\mcal D(x) \geq \mcal D(y)$.} distribution $\mathcal D$ with decision tree complexity $\ell$, there exists an algorithm $\mathcal A'$ that is a PAC learner for $\mathscr C$ that is $(m \ell)^\ell$-sample replicable under $\mathcal{ D}$ and runs in time $\poly((m \ell)^\ell)$.
\end{thm}

This implies for instance that for $\ell = O(1)$, the blowup is polynomial in the runtime of the uniform learner while if $\ell = O(\log d)$, the running time is quasi-polynomial in $d$.
We note that our result can also be extended to general (non-monotone) distributions $\mcal D$ if we assume access to a \emph{(subcube) conditional sampling oracle}. For 
formal details, we refer the reader to \Cref{sec:lifting-main}. 

As an application of this framework, we show how to black-box transform replicable learners
for parities that work under the uniform distribution to replicable learners that work
under some other unknown distribution $\mathcal D$, 
where the sample complexity and running time
of the transformation depends on the decision tree complexity of $\mathcal{D}$. This result is hence related to the question of \citet{impagliazzo2022reproducibility} since we can design distributions $\mcal D$ over $\{0,1\}^d$ with small decision tree complexity for which Gaussian elimination fails to be replicable.
We refer to \Cref{sec:main-parity} for the formal statement.

\begin{cor}
[Informal, see \Cref{thm:parities} and \Cref{thm:replicable parities}] 
\label{thm:main3}
Let $\mcal X = \{0,1\}^d$ and $m = \poly(d)$. Then, for any
(unknown) monotone distribution $\mcal D$ with decision tree complexity $\ell$, there exists a PAC learner for parities that is $(m \ell)^{\ell}$-sample replicable under $\mcal D$ and runs in time $\poly((m \ell)^{\ell})$.

Moreover, for any $m' = m^{\Theta(1)}$, there exists some monotone distribution $\mcal D$ over $\{0,1\}^d$ with decision tree complexity $\ell = \Theta(1)$ so that Gaussian elimination, with input
$m'$ labeled examples,
fails to PAC learn parities replicably under $\mcal D$ with constant probability.
\end{cor}

\paragraph{Replicability \& Privacy.}
Returning to the works of \citet{ghazi2021user,bun2023stability,kalavasis2023statistical}, it is known that replicable PAC learning is statistically equivalent to approximate differentially private PAC learning. 
We hence ask:
\begin{center}
Q4. \emph{How does replicability computationally relate to private learning?}
\end{center}
The recent work of \citet{bun2023stability} provided a computational separation between \emph{approximate} differential privacy and replicability. In particular, they showed that a replicable learner can be efficiently transformed into an approximately differentially private one, but the other direction is hard under standard cryptographic assumptions.
To be more specific, they provided a concept class that is efficiently learnable by an approximate DP learner, 
but not efficiently learnable by a replicable learner,
assuming the existence of one-way functions.
Based on this hardness result, we ask whether we can transform a \emph{pure} DP algorithm into a replicable one.  
We show the following result.
\begin{thm}
[Informal; see \Cref{thm:efficient-pure-dp-to-replicable}]
\label{thm:main5-intro}
 Let $\mcal D_{XY}$ be a distribution
    on $\mcal X \times \{0,1\}$ that is realizable with respect to some 
    concept class $\mscr C$.
    Let $\mcal A$ be an efficient pure DP learner for $\mscr C$.
    Then,
    there is a replicable learner $\mcal A'$
    for $\mscr C$ that runs in time polynomial
    with respect to the error, confidence, and
    replicability parameters 
    but exponential in the representation dimension\footnote{The \emph{representation dimension} is a combinatorial dimension, similar to VC dimension, that characterizes which classes are PAC learnable by pure DP algorithms \citep{beimel2013characterizing}.}
    of $\mscr C.$
\end{thm}

{We reiterate that this transformation is efficient with respect to the
correctness, confidence and replicability parameters $\alpha, \beta, \rho$, but \emph{it 
could be inefficient} with respect to some parameter that captures 
the complexity of the underlying concept class $\mscr C$ (such as the representation dimension of the class, e.g., the margin parameter in the case of large-margin halfspaces).
To the best of our knowledge, the same holds for the transformation
of a pure DP learner to an online learner of \citet{gonen2019private} and it would be interesting to show that this is unavoidable.}

\paragraph{The Computational Landscape of Stability.}  Our work studies several computational aspects of replicability and its connections with other stability notions such as online learning, SQ learning, and differential privacy.
Combining our results with the prior works of \citet{blum2005practical,gonen2019private,ghazi2021user,impagliazzo2022reproducibility,bun2023stability,kalavasis2023statistical} yields the current computational landscape of stability depicted in \Cref{fig:landscape}.
It is a byproduct of the following results connecting (i) replicability, (ii) approximate DP, (iii) pure DP, (iv) online learning, and, (v) statistical queries.

\newcommand{\computationalLandscapeFigure}{
\begin{tikzcd}[scale=1.2,ampersand replacement=\&,cramped]
    \&\&\&\&\&\&\&\&\& {\varepsilon\text{-DP}} \\
    \&\&\& {\text{Replicability}} \&\&\&\& {\text{Online}} \\
    \\
    \\
    {\text{SQ}} \&\&\&\&\&\&\& {(\varepsilon, \delta)\text{-DP}} \\
    \\
    \&\&\&\& {}
    \arrow["{\text{\color{black}\citep{blum2005practical}}}"{description}, color={rgb,255:red,26;green,102;blue,26}, curve={height=30pt}, Rightarrow, from=5-1, to=5-8]
    \arrow["{\text{\color{black}\citep{gonen2019private}}}"{description}, color={rgb,255:red,235;green,192;blue,52}, Rightarrow, from=1-10, to=2-8, dashed]
    \arrow["{\text{\color{black}By definition}}", color={rgb,255:red,26;green,102;blue,26}, curve={height=-30pt}, Rightarrow, from=1-10, to=5-8]
    \arrow["{\text{\color{black}This work, \Cref{thm:main1}}}", "|"{marking, text={rgb,255:red,163;green,41;blue,41}}, color={rgb,255:red,163;green,41;blue,41}, from=2-4, to=2-8]
    \arrow["\substack{\text{\color{black} \citep{impagliazzo2022reproducibility},}\\ \text{\color{black}This work, \Cref{thm:main3}}}"{description, pos=0.6}, "|"{marking, text={rgb,255:red,163;green,41;blue,41}, pos=0.3}, color={rgb,255:red,163;green,41;blue,41}, curve={height=-30pt}, from=2-4, to=5-1]
    \arrow["{\text{\color{black}\citep{bun2020computational}}}"{description, pos=0.3}, "|"{marking, text={rgb,255:red,163;green,41;blue,41}}, color={rgb,255:red,163;green,41;blue,41}, curve={height=33pt}, from=5-8, to=2-8]
    \arrow["{\text{\color{black}\citep{bun2024private}}}"{description, pos=0.3}, "|"{marking, text={rgb,255:red,163;green,41;blue,41}}, color={rgb,255:red,163;green,41;blue,41}, curve={height=10pt}, from=2-8, to=5-8]
    \arrow["{\text{\color{black}This work, \Cref{thm:efficient-pure-dp-to-replicable}}}"{description}, color={rgb,255:red,235;green,192;blue,52}, curve={height=30pt}, Rightarrow, from=1-10, to=2-4, dashed]
    \arrow["\substack{\text{\color{black}[\citealp{ghazi2021user};} \\ \text{\color{black}\citealp{bun2023stability};} \\ \text{\color{black}\citealp{kalavasis2023statistical}]}}"{description, pos=0.5}, color={rgb,255:red,26;green,102;blue,26}, curve={height=-18pt}, Rightarrow, from=2-4, to=5-8]
    \arrow["{\text{\color{black}\citep{bun2023stability}}}"{description, pos=0.4}, 
        color={rgb,255:red,235;green,192;blue,52},
        curve={height=-30pt}, 
        Rightarrow,
        from=5-8, 
        to=2-4,
        dashed]
    \arrow["{\text{\color{black}\citep{impagliazzo2022reproducibility}}}"{description}, color={rgb,255:red,26;green,102;blue,26}, curve={height=-30pt}, Rightarrow, from=5-1, to=2-4]
\end{tikzcd}
}

\ifarxiv
\begin{sidewaysfigure}
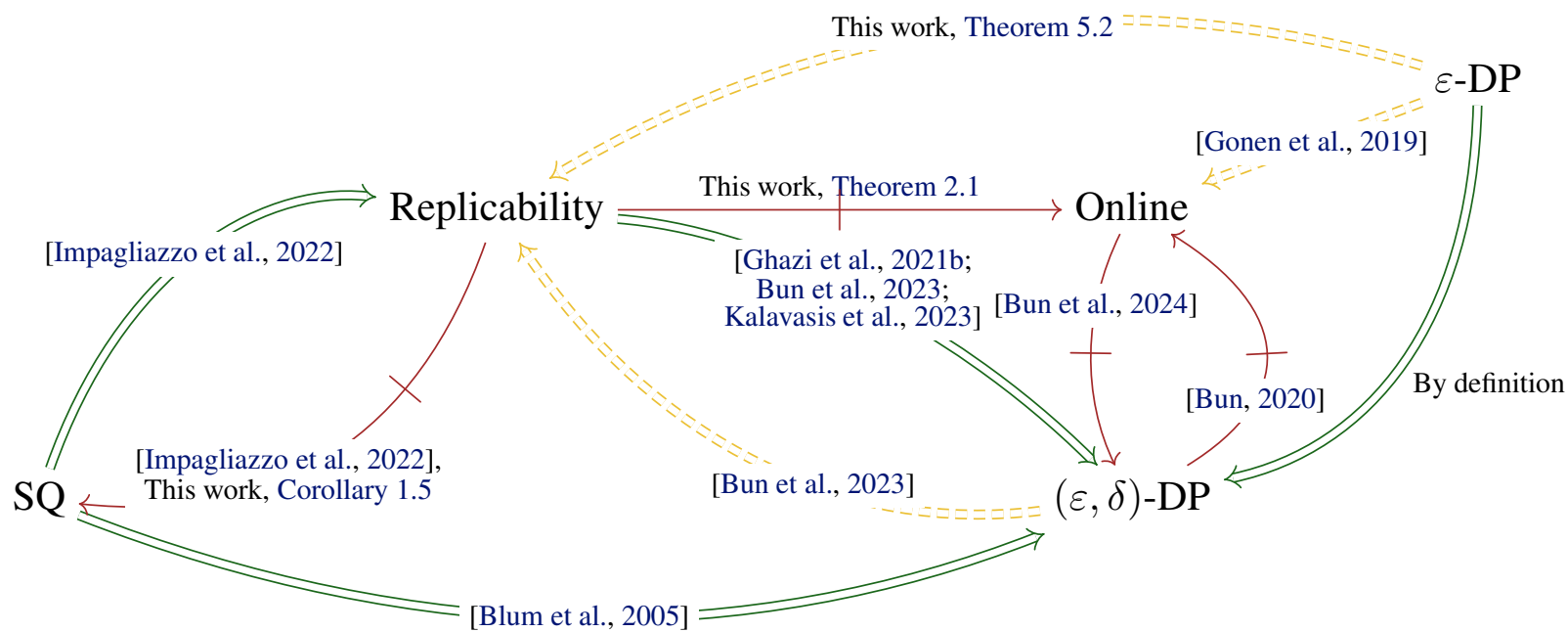

\else
\begin{figure}[htbp!]
\fi
    \centering
    \adjustbox{scale=1.6,center}{\computationalLandscapeFigure}
    \caption{The computational landscape of stability.
    A green double arrow ($\Rightarrow$) from tail to head
    indicates that an efficient learner for a task in the setting of the arrow tail
    can be black-box transformed into an efficient learner for the same task in the setting of the arrow head.
    Meanwhile,
    an orange dashed double arrow ($==\succ$) from tail to head indicates that an efficient learner for a task in the tail setting
    can be black-box transformed into an efficient learner for the same task in the head setting,
    under some additional assumptions.
    Finally,
    a red slashed single arrow ($\nrightarrow$) from tail to head
    indicates that there is a learning task for which an efficient learner exists in the setting of the arrow tail
    but no efficient learner can exist in the setting of the arrow head,
    possibly under some cryptographic assumptions.\\
    }\label{fig:landscape}
\ifarxiv
\end{sidewaysfigure}
\else
\end{figure}
\fi

\paragraph{Black-Box Transformations.}
\begin{enumerate}[itemsep=0em,leftmargin=*,topsep=0pt]
    \item Pure DP can be efficiently\footnote{The transformation is efficient with respect to the correctness and confidence
parameters but could be inefficient with respect to other parameters of the class such as the representation dimension.} transformed to Online \citep{gonen2019private}.
    \item Replicability can be efficiently transformed to Approximate DP \citep{ghazi2021user,bun2023stability,kalavasis2023statistical}.
    \item SQ can be efficiently transformed to Approximate DP
    \citep{blum2005practical}.\footnote{The reduction is efficient as it simply requires adding Gaussian noise to each statistical query. However, it is impractical for most tasks of interest as the resulting privacy guarantees are relatively weak.}
    \item SQ can be efficiently transformed to Replicable \citep{impagliazzo2022reproducibility}.
    \item Pure DP can be efficiently\footnote{The transformation is efficient with respect to the correctness and confidence
parameters but could be inefficient with respect to other parameters of the class.}  transformed to Replicable ({\color{black} this work}, \Cref{thm:efficient-pure-dp-to-replicable}).
\end{enumerate}

\paragraph{Caveats of Transformations.}
The transformations from pure $\varepsilon$-DP learners to replicable and online learners
may incur exponential computation time in the \emph{representation dimension} \citep{beimel2013characterizing} of the underlying hypothesis class.

Regarding the approximate DP reduction to replicability:
The transformation provided by \citet{bun2023stability} uses correlated sampling, 
which necessitates that the output space
of the algorithm is finite. 
To be more precise, based on \citep{bun2023stability}, 
there is an efficient transformation from approximate DP to perfectly generalizing algorithms. 
Next, the authors use correlated sampling to obtain a replicable learner as follows.
Given a perfectly generalizing algorithm $A$ and sample $S$,
the correlated sampling strategy is applied to the distribution of outputs of 
$A(S)$. 
Hence, the output space of $A$ should be finite.
In the PAC learning setting,
to ensure that the algorithm
has finite output space,
one sufficient 
condition is that the domain $\mathcal{X}$
is finite. 
The correlated sampling step can
be explicitly implemented via rejection sampling from the output space of $A$.
The acceptance probability is controlled
by the probability mass function of $A(S)$. 
As a result, in general, it is not computationally efficient.
For instance, if the finite input space to the correlated sampling strategy is $\{0,1\}^d$, then the runtime of the algorithm could be $\exp(d)$, 
since the acceptance probability is exponentially small in the dimension in the worst case. 

For the specific case of PAC learning,
there is another transformation from approximate DP
to replicability that holds for 
countable domains $\mathcal{X}$ that was 
proposed by
\citet{kalavasis2023statistical}, but this approach goes
through the Littlestone dimension of the
class and might not even be computable
in its general form.

\paragraph{Separations.}
There is a concept class that can be learned efficiently
\begin{enumerate}[itemsep=0em,leftmargin=*,topsep=0pt]
    \item by a replicable PAC learner ({\color{black} this work}, \Cref{thm:main1}),
    but not an efficient online one under OWF \citep{blum1994separating};
    \item by an approximate DP PAC learner,
    but not an efficient online one under OWF \citep{bun2020computational};
    \item by an online learner,
    but not an efficient approximate DP PAC one under cryptographic assumptions\footnote{The specific assumptions are technical and we omit the specific statements.
    Roughly, the assumptions lead to the possibility to build indistinguishability obfuscation for all circuits.} \citep{bun2024private};
    \item by a replicable PAC learner under uniform marginals (\cite{impagliazzo2022reproducibility}) or more general marginals ({\color{black} this work}, \Cref{thm:main3}),
    but not an efficient SQ one;
    \item by an approximate DP PAC learner,
    but not an efficient replicable one under OWF \citep{bun2023stability}.
\end{enumerate}

\subsection{Related Work}
\textbf{Replicability.}
Pioneered by \citet{impagliazzo2022reproducibility},
there has been a growing interest from the learning theory community in studying
replicability as an algorithmic property. 
\citet{esfandiari2023replicableb, esfandiari2023replicable} studied 
replicable algorithms in the context of multi-armed bandits and clustering. Later, \citet{eaton2023replicable, karbasi2023replicability} studied replicability 
in the context of Reinforcement Learning (RL) and 
designed algorithms that achieve various notions 
of replicability.
Recently, \citet{bun2023stability} established statistical equivalences and separations between replicability
and other notions of algorithmic stability such as differential privacy
when the domain of the learning problem is finite and provided some computational
and statistical
hardness results to obtain these equivalences, under cryptographic assumptions. 
Subsequently, \citet{kalavasis2023statistical} proposed a relaxation of the replicability
definition of \citet{impagliazzo2022reproducibility}, 
showed its statistical
equivalence to the notion of replicability for countable domains\footnote{We remark that this equivalence for finite domains can also be obtained, implicitly, from the results of \citet{bun2023stability}. Connections between replicability and the Littlestone dimension go back to \citet{ghazi2021user}.}
and extended some of the equivalences from \citet{bun2023stability} to countable
domains.
\citet{chase2023replicability, dixon2023list} proposed
a notion of \emph{list-replicability},
where the randomness is not shared across the executions of the algorithm, but the outputs
are required to belong to a list of 
small cardinality instead of being identical.
Both of these works developed algorithms
that work in the
\emph{realizable} setting, and later \citet{chase2023local} showed 
that, surprisingly, it is impossible to 
design list-replicable learning algorithms
for infinite classes in the \emph{agnostic} setting.
Recently, \citet{moran2023bayesian} established even more
statistical
connections between replicability and other notions
of stability, by dividing them into two categories
consisting of distribution-dependent and distribution-independent
definitions. Recent work by \cite{kalavasis2024replicable} studies replicable algorithms for large-margin halfspaces.

\paragraph{Computational Separations \& Transformations in Stability.} 
The seminal result of \citet{blum1994separating} illustrated a computational separation between PAC and online learning. 
Later, \citet{bun2020computational} obtained a similar separation between private PAC and online learning. More recently, \citet{bun2023stability} showed that there exists a concept class that admits an efficient approximate DP learner but not an efficient replicable one, assuming the existence of OWFs. 
Contributing to this line of work, our \Cref{thm:main1} is of similar flavor.

Shifting our attention to SQ learning, there is an intuitive similarity between learning from noisy examples and private learning: algorithms for both problems must be robust to
small variations in the data (see also \citet{blum2005practical}). 
Parities are a canonical SQ-hard problem, meaning that
no efficient SQ algorithm for this class exists. 
\citet{kasiviswanathan2011can} designed
an efficient private learner for learning parities,
which dispels the similarity between learning with noise and private learning. \citet{georgiev2022privacy} showed that while polynomial-time private algorithms for learning
parities are known, the failure probability of any such
algorithm must be larger than what can be achieved in exponential 
time, or else $\mathrm{NP} = \mathrm{RP}$. 
{Finally, recent work by \cite{bun2024private} gives a concept class that admits an online learner running in polynomial time
with a polynomial mistake bound, but for which there is no computationally efficient {approximate} differentially private PAC learner, under cryptographic assumptions.}

Moving on to transformations, our \Cref{thm:main4} builds upon the result of \citet{blanc2023lifting} who designed a framework 
that transforms computationally efficient algorithms
under uniform marginal distributions, to algorithms
that work under some other distribution, where 
the complexity of the transformation scales with some
particular notion of distance between the two distributions.
Essentially, our result can be viewed as a replicable
framework of the same flavor, with a small additional computational and
statistical overhead that scales with the replicability parameter. 
Finally, 
our approach to transform a pure DP learner into a replicable learner (cf. \Cref{thm:main5-intro}) is inspired by \citet{gonen2019private},
who provided a transformation from pure DP learners to online learners.

\subsection{Notation}
In general,
{we use $\mcal A$ to denote an algorithm}.
For unsupervised problems, we usually denote by $\mcal D$ the distribution over input examples. In the case of supervised problems,
{we use $\mscr C$ to denote the concept class in question},
$\mcal D$ to denote the marginal distribution of the feature domain $\mcal X$,
and $\mcal D_{\mcal X \mcal Y}$ to refer to the joint distribution over labeled examples $(x,y) \in \mcal X \times \mcal Y$. 
Throughout our work, we use $\alpha, \beta$ to
refer to the error and failure probability parameters of the algorithm,
$\varepsilon, \delta$ to denote the approximate DP parameters, and
$\rho$ for the replicability parameter. In most cases,
the feature domain $\mcal X$ is a subset of a high-dimensional space
and we use $d$ to denote the dimension of that space.

\section{Efficient Replicability and Online Learning}
\label{sec:online}
Our first main result shows that replicability and online
learning are \emph{not} computationally equivalent, assuming the existence
of one-way functions. 
This hardness result 
is based on a construction from \citet{blum1994separating},
who defined a concept class denoted by $\OWS$,
which is efficiently PAC learnable but not online learnable in polynomial time,
assuming the existence of one-way functions.
Blum's construction builds upon the Goldreich-Goldwasser-Micali pseudorandom function generator \citep{goldreich1986construct}
to define families of ``one-way'' labeled sequences $(\sigma_1, b_1), \dots, (\sigma_r, b_r)\in \Set{0, 1}^d\times \Set{0, 1},$
for some $r = \omega(\poly(d))$.
These string-label pairs can be efficiently computed in the forward direction,
but are hard to compute in the reverse direction.
Specifically,
for any $i<j$,
it is easy to compute $\sigma_j, b_j$ given $\sigma_i$.
On the other hand,
it is hard to compute $\sigma_i, b_i$ given $\sigma_j$.

The essence of the difficulty in the online setting is that
an adversary can present to the learner the sequence in reverse order,
i.e., $\sigma_r, \sigma_{r-1}, \dots,\sigma_1$.
Then, the labels $b_i$ are not predictable by a polynomial time learner.
However, in the PAC setting, for any distribution over 
the sequence $\{(\sigma_i,b_i)\}_{i \in [r]}$,
a learner which is given $n$ labeled examples can identify the string $\sigma_{i^*}$ with smallest index $i^*$ in the sample.
Then, it can perfectly and efficiently predict the label of any string that comes after it. This approximately amounts to a $(1-\nicefrac1n)$ fraction of the underlying population.

It is not hard to see that the PAC learner for \OWS is not replicable,
since the minimum index of a sample can vary wildly between samples.
The high-level idea of our approach to making the algorithm replicable 
is to show that we can replicably identify an \emph{approximate} minimum index
and output the hypothesis that (efficiently) forward computes from that index. Our main result,
proven in \Cref{sec:replicable but not online},
is as follows.

\begin{restatable}{thm}{mainTheoremOne}
\label{thm:main1}
Let $d \in \mathbb N$.
The following hold:
\begin{enumerate}[(a),leftmargin=*,topsep=0pt]
    \item \citep{bun2020computational}~Assuming the existence of one-way functions, the concept class $\OWS$ with input domain $\{0,1\}^d$ cannot be learned by an efficient no-regret algorithm, i.e., an algorithm that for some $\eta > 0$
    achieves expected regret $\E[R_T] \leq \poly(d) \cdot T^{1-\eta}$ using time $\poly(d,T)$ in every iteration.\footnote{{We remark that this stronger requirement in regret is necessary since the trivial random guessing algorithm achieves $o(T)$ regret in finite domains (cf. \Cref{sec:online preliminary}).}}
    
    \item (\Cref{thm:replicable learner OWS}) The concept class $\OWS$ with input domain $\{0,1\}^d$ can be learned by a $\rho$-replicable $(\alpha,\beta)$-PAC learner with sample complexity $m = \poly(d, \nicefrac1\alpha, \nicefrac1\rho, \log(\nicefrac1\beta))$ and $\poly(m)$ running time.
\end{enumerate}
\end{restatable}

\section{Lifting Replicable Uniform Learners}
\label{sec:lift}
\label{sec:lifting-main}
In this section, we present our replicable lifting framework in further detail. First, we will need the following technical definition.
\begin{defn}
[Closed under Restrictions]
\label{def:restrictions}
    A concept class $\mathscr C$ of functions $f : \{0,1\}^d \to \{0,1\}$ is closed under restrictions if, for any $f \in \mathscr C, i \in [d]$ and $b \in \{0,1\}$, the restriction
    $f_{i = b}$ remains in $\mathscr C$, 
    where $f_{i = b}(x) = f(x_1,...,x_{i-1},b,x_{i+1},...,x_d)$ for any $x \in \{0,1\}^d$. 
\end{defn}
Our replicable lifting result works for concept classes that satisfy the closedness under restrictions property.
Also, recall that a distribution $\mcal D$ over $\{0,1\}^d$ is \emph{monotone} (\Cref{def:monotone}) if whenever $x \succeq y$ ($x$ is greater than $y$ in the partial ordering of the poset), it holds $\mcal D(x) \geq \mcal D(y)$.
Our algorithm for lifting replicable uniform learners to replicable learners under some unknown distribution $\mcal D$ is presented in \Cref{thm:lift}.
This lifting approach is inspired by the work of \citet{blanc2023lifting}, who designed the non-replicable variant of this transformation.
We note that our algorithm, similar to the algorithm of \citet{blanc2023lifting},
only requires sample access to $\mcal D$ for monotone distributions, while,
for arbitrary non-monotone probability measures, 
our algorithm requires access to a \emph{conditional sampling oracle} (cf. \Cref{def:conditional sampler}).
Our replicable lifting theorem reads as follows.

\begin{restatable}
[Lifting Replicable Uniform Learners]
{thm}
{rLiftUniformLearners}
\label{thm:lift} 
    Consider a concept class $\mathscr{C}$ of functions $f: \{0,1\}^d \to \{0,1\}$ closed under restrictions. 
    Suppose we are given black-box access to an algorithm such that for any $\alpha', \rho', \beta' \in (0,1)$,
    given $\poly(d, \nicefrac1{\alpha'}, \nicefrac1{\rho'}, \log(\nicefrac1{\beta'}))$ samples from $\mcal U$,
    \begin{enumerate}[(i),leftmargin=*,topsep=0pt]
        \item is $\rho'$-replicable with respect to the uniform distribution $\mcal U$,
        \item PAC learns $\mathscr{C}$ under the uniform distribution to accuracy $\alpha'$ and confidence $\beta'$, and,
        \item terminates in time $\poly(d, \nicefrac1{\alpha'}, \nicefrac1{\rho'}, \log(\nicefrac1{\beta'})).$
    \end{enumerate}
    Let $\alpha, \rho\in (0, 1)$ and $\beta\in (0, \nicefrac\rho3)$.
    For $m = \poly(d, \nicefrac1\alpha, \nicefrac1\rho, \log(\nicefrac1\beta))$ and
    $
        M 
        = {\poly(d, \nicefrac1\alpha, \nicefrac1\rho, \log(\nicefrac1\beta))^{O(\ell)}}
    $,
    the following cases hold:

    \begin{enumerate}[(a),leftmargin=*,topsep=0pt]
        \item If $\mathcal D$ is a monotone distribution over $\{0,1\}^d$ representable by a depth-$\ell$ decision tree, there is an algorithm that draws $M$ samples from $\mcal D$, is $\rho$-replicable with respect to $\mathcal D$,
        PAC learns $\mathscr{C}$ under $\mathcal D$ with accuracy $\alpha$ and confidence $\beta$,
        and terminates in $\poly(M)$ time.
        \item If $\mathcal D$ is an arbitrary distribution over $\{0,1\}^d$
        representable by a depth-$\ell$ decision tree, there is an algorithm that draws $M$ labeled examples as well as $M$ conditional samples (cf. \Cref{def:conditional sampler}) from $\mathcal D$, is $\rho$-replicable with respect to $\mathcal D$,
        PAC learns $\mathscr{C}$ with respect to $\mathcal D$ with accuracy $\alpha$ and confidence $\beta$,
        and terminates in $\poly(M)$ time.
    \end{enumerate}
\end{restatable}

The high-level idea of the reduction proceeds as follows. 
Let us consider the realizable PAC setting, i.e., the labels are consistent with some $f^\star \in \mathscr C$.
Let us assume black-box access to a replicable uniform learner $\mcal A_\mathscr C^{\mathcal U}$ for $\mathscr C$ and access to i.i.d. samples of the form $(x, f^\star(x))$, where $x \sim \mathcal D$. 
Our goal is to (efficiently) obtain an algorithm $\mcal A_{\mathscr C}^\mathcal D$ that achieves small misclassification error with respect to the unknown distribution $\mathcal D$ and target $f^\star$ and is also replicable under $\mcal D$.
The promise is that $\mcal D$ has a decision tree representation of depth $\ell$.\footnote{{By performing an iterated doubling if necessary, we may assume without loss of generality that $\ell$ is known.}}

The first step is to draw enough samples $(x, f^\star(x)), x \sim \mathcal D$, and compute the decision tree representation of $\mathcal D$. This is an unsupervised learning task and it only uses the feature vectors of the training set. We design a replicable algorithm for this step (cf. \Cref{thm:learn-DT}), which could be of independent interest.
The next observation is crucial:
any discrete distribution on $\{0,1\}^d$ can be expressed as a mixture of uniform distributions conditioned on non-overlapping sub-cubes. To see this, notice that any root-to-leaf path in the estimated decision tree representation corresponds to a sub-cube of $\{0,1\}^d$ and, conditioned on this path, the remaining coordinates follow a uniform law.
Hence, after an appropriate re-sampling procedure, one can employ the black-box replicable learner $\mcal A_\mathscr C^\mathcal U$ to any one of the leaves $t$ of the tree decomposition of $\mathcal D$ and obtain a classifier $f_t$. For this step, the fact that $\mathscr C$ is closed under restrictions is crucial. 
Intuitively, 
if we wish to implement this idea in a replicable manner
and need overall replicability parameter $\rho$,
it suffices to use the uniform replicable algorithm in each leaf with parameter $O(\nicefrac\rho{2^\ell})$
since we make at most $2^\ell$ calls to the uniform PAC learner (the decision tree complexity of the target is $\ell$). Finally, given a test example $x \sim \mathcal D$,
one computes the leaf $t$ that corresponds to the sub-cube that $x$ falls into and uses the (replicable) output $f_t$ of the associated uniform
PAC learner to guess the correct label.

For the formal analysis, we refer to \Cref{sec:lifting} and \Cref{apx:lifting}.

\section{Efficient Replicability and SQ Learning: Parities}
\label{sec:main-parity}
In this section, we provide an application of the general lifting framework we described
in \Cref{sec:lifting-main}.
One of the main results of the seminal work of \citet{impagliazzo2022reproducibility} is that any SQ algorithm can be made replicable. This result allows a great collection of tasks to be solved replicably since the SQ framework is known to be highly expressive \citep{kearns1998efficient}.

The primary motivation of this section comes from the question of \citet{impagliazzo2022reproducibility} on whether the class of parities can be PAC learned by an efficient replicable algorithm when the marginal distribution $\mcal D$ is not uniform over $\{0,1\}^d$.
Let us define our concept class of interest.\footnote{We consider a superset of parities which we call \emph{affine parities} as we would like our concept class to be closed under restrictions (cf. \Cref{def:restrictions}).}

\begin{defn}[Affine Parities]
    Let $S \subseteq [d]$ be some subset of $[d]$
    and $b \in \{0,1\}$ a bias term.
    Define $f_{S,b} : \{0,1\}^d \to \{0,1\}$ to be the biased parity of the bits in $S$, 
    namely $f_{S,b}(x) = b + \sum_{i \in S} x_i$. 
    The concept class of \emph{affine parities} over $\{0,1\}^d$ is the set $\mathscr{C} = \{ f_{S,b} : S \subseteq [d], b \in \{0,1\}\}$.
    For any distribution $\mathcal D$ over $\{0,1\}^d$, we consider the supervised learning problem $\mathsf{AffParity}$, where the learner observes i.i.d. samples of the form $(x, f^\star(x))$, where $x \sim \mathcal D$ and $f^\star \in \mathscr C.$
\end{defn}

As observed by \citet{impagliazzo2022reproducibility}, there is a replicable algorithm for PAC learning parities (i.e., the subclass $\mathsf{Parity}$ obtained by setting $b=0$ in $\mathsf{AffParity}$) under the uniform distribution: draw roughly $O(d)$ samples so that with high probability the dataset will contain a basis. Then, in two distinct executions, the sets that standard Gaussian elimination outputs will be the same. A similar approach works for the class of affine parities and is described below.

\begin{restatable}{lem}{rAffineParity}\label{lem:replicable affine parity}
    The concept class of affine parities
    $\mathsf{AffParity}$ over $\{0,1\}^d$ admits a $\rho$-replicable algorithm that perfectly learns any concept with respect to the uniform distribution $\mcal U$ with probability of success at least $1-\beta$.
    The algorithm has $O(\poly(d, \log\nicefrac1{\rho\beta}))$ sample and time complexity.
\end{restatable}

The algorithm that attains the guarantees of \Cref{lem:replicable affine parity} is a simple adaptation of the Gaussian elimination for learning standard parities.
We provide the pseudocode in \Cref{alg:replicable affine parity}
and defer the proof of correctness to \Cref{apx:replicable affine parity}.

\paragraph{Application of \Cref{thm:lift}}
Since the class of affine parities over $\{0,1\}^d$ is closed under restrictions (cf. \Cref{lem:closed}) and there is a learner for this class {under uniform marginals} that is replicable and efficient (cf. \Cref{lem:replicable affine parity}), we can obtain an algorithm that replicably PAC learns the class $\mathsf{AffParity}$ under more general distributions. In particular, the following result is an immediate application of our lifting framework. 

\begin{cor}
\label{thm:parities}
    Let $\alpha, \rho\in (0, 1)$ and $\beta\in (0, \nicefrac\rho3)$.
    Let $\mcal X = \{0,1\}^d$.
    For
    {$M = \poly(d, \nicefrac1\alpha, \nicefrac1\rho, \nicefrac1\beta)^{O(\ell)}$},
    the following cases hold:
    \begin{itemize}[(a),leftmargin=*,topsep=0pt,itemsep=0pt]
        \item For any
        (unknown) monotone distribution $\mcal D$ over $\mcal X$ with decision tree complexity $\ell$, there exists an algorithm that is a $\rho$-replicable learner for the concept class $\mathsf{AffParity}$
        under $\mcal D$, and requires $M$ samples and
     running time $\poly(M)$
        to get accuracy $\alpha$ and confidence $\beta$.
        
        \item For any distribution $\mcal D$ over $\mcal X$ with decision tree complexity $\ell$, there exists an algorithm that is a $\rho$-replicable learner for the concept class $\mathsf{AffParity}$
        under $\mcal D,$ and requires $M$ labeled examples, $M$ conditional samples from $\mcal D$
        and running time $\poly(M)$
        to get accuracy $\alpha$ and confidence $\beta.$
    \end{itemize}
\end{cor}

\paragraph{Back to the Question of \citet{impagliazzo2022reproducibility}.}
\citet{impagliazzo2022reproducibility} raised the question of when parities over $\{0,1\}^d$ can be efficiently PAC learned by a replicable algorithm under some marginal distribution $\mcal D$ that is not uniform or, more broadly, that causes Gaussian elimination to be non-replicable.
Our \Cref{thm:parities} makes progress on this question. First, we note that in our setting, we do not require knowledge of $\mcal D$. 
Second, one can design examples of (monotone) distributions for which our lifting framework produces a replicable polynomial time PAC learner but naive Gaussian elimination\footnote{We note that the examples we design are difficult instances for the standard Gaussian elimination algorithm but some pre-processing of the dataset (e.g., deleting some constant fraction of samples) could potentially make Gaussian elimination replicable.} fails to be replicable with constant probability (cf. \Cref{thm:replicable parities}). 
Note that even PAC learning parities for the distribution described in \Cref{thm:replicable parities} is SQ-hard. 
We believe that our lifting framework can be seen as a systematic way of bypassing some instabilities arising from the algebraic structure of standard Gaussian elimination.

 As a final remark, we note that one could potentially design much more complicated (still monotone) distributions than the one of \Cref{thm:replicable parities}, where even various adaptations of Gaussian elimination (e.g., pre-processing of the dataset, data deletion) would fail to be replicable but our lifting framework would guarantee replicability (and efficiency, provided small decision tree complexity). On the other hand, it is not evident whether parity learning remains SQ-hard under these ``harder'' distributions.
 
\section{Efficient Replicability and Private Learning}

In this section, we study connections between efficient DP 
learnability and efficient replicable learnability of a concept
class. 
As we mentioned in the introduction, \citet{bun2023stability} and subsequently 
\citet{kalavasis2023statistical} established a \emph{statistical}
equivalence between approximate DP learnability
and replicable learnability of a concept class when the domain is
finite \citep{bun2023stability} or countable \citep{kalavasis2023statistical}. Moreover,
\citet{bun2023stability} showed that this equivalence
does not hold when one takes into account the computational complexity
of these tasks.
\begin{prop}
[Section 4.1 in \citet{bun2023stability}]
There exists a class that is efficiently PAC learnable by an approximate DP algorithm but, assuming one-way functions exist, it cannot be learned efficiently by a replicable algorithm.
\end{prop}
We remark that there is an efficient converse transformation, i.e., a computationally efficient transformation from a replicable
learner to an approximate DP learner \citep{bun2023stability}.
Inspired by \citet{gonen2019private}, we ask whether one can transform a \emph{pure} DP learner to a replicable one. 
Our main result, 
proven in \Cref{apx:private to replicable},
is the following.
\begin{restatable}[From Pure DP Learner to Replicable Learner]{thm}{dpToReplicable}\label{thm:efficient-pure-dp-to-replicable}
    Let $\mcal X$ be some input domain, $ \mcal Y = \{0,1\}$,
    and $\mcal D_{\mcal X\mcal Y}$ be a distribution
    on $\mcal X \times \mcal Y$ that is realizable with respect to some 
    concept class $\mathscr{C}$. 
    Let $\mcal A$ be a pure DP learner that, 
    for any $\alpha, \varepsilon, \beta \in (0,1)$,
    needs $m(\alpha, \varepsilon, \beta, \mscr C) = \poly(\nicefrac1{\alpha}, \nicefrac1{\varepsilon}, \log(\nicefrac1{\beta}), \mathsf{dim}(\mscr C))$ i.i.d. samples from $\mcal D_{\mcal X \mcal Y}$ 
    and $\poly(m)$ running time
    to output a hypothesis that has error at most $\alpha$, 
    with probability $1-\beta$ in an $\varepsilon$-DP way.
    Then, for any $\alpha', \rho, \beta' \in (0,1)$
    there is a $\rho$-replicable learner $\mcal A'$
    that outputs a hypothesis with error at most $\alpha'$ with probability at least $1-\beta'$ and requires $\poly(\nicefrac1{\alpha'}, \nicefrac1\rho, \log(\nicefrac1{\beta'}), \mathsf{dim}(\mathscr C))$ i.i.d.
    samples from $\mcal D_{XY}$ and $\poly(\nicefrac1{\alpha'}, \nicefrac1\rho, \log(\nicefrac1{\beta'})) \cdot \exp(\mathsf{dim}(\mscr C))$ running time.
\end{restatable}

{In the above, we denote by $\mathsf{dim}(\mscr C)$ some dimension that describes the complexity of the concept class $\mscr C$ that arises in the sample complexity of our pure DP learner. 
A natural candidate is the \emph{representation dimension} \citep{kasiviswanathan2011can}.
As we alluded to 
before, this transformation is efficient with respect to 
the parameters $\alpha, \beta, \rho.$ On the other side, the sample complexity is polynomial in the representation dimension but the running time
is exponential. We leave as an open question if it is possible to avoid this dependence.}

We remark that in principle, one could use the reduction from \citet{bun2023stability}. 
The catch is that this reduction is based on correlated sampling so it requires 
i) the output space of the algorithm to be finite and 
ii) even under finite output spaces, it needs exponential time in the size of that space.

\section{Conclusion}\label{sec:conclusion}
In this work, we have studied the computational aspects
of replicability and several connections to other important notions
in learning theory including online learning, SQ learning, and DP PAC
learning. We believe that there are several interesting questions left open from our work. 
First, it would be interesting to see
if there is a 
computationally efficient transformation from online learners to 
replicable learners. Then, it would be important to derive
replicable learners from pure DP learners which are efficient 
with respect to the complexity of the underlying 
concept class. Regarding parities, it is still open whether
we can design efficient replicable algorithms for $\emph{every}$
distribution $\mcal D$.

\section*{Acknowledgments}
Alkis Kalavasis was supported by the Institute for Foundations of Data Science at Yale. Amin Karbasi acknowledges funding in direct support of this work from NSF (IIS-1845032), ONR (N00014- 19-1-2406), and the AI Institute for Learning-Enabled Optimization at Scale (TILOS). Grigoris Velegkas was supported in part
by the AI Institute for Learning-Enabled Optimization at Scale (TILOS). 
Felix Zhou acknowledges the support of the Natural Sciences and Engineering Research Council of Canada (NSERC). 
\begingroup
\sloppy
\bibliography{references}

\begin{thebibliography}{49}
\providecommand{\natexlab}[1]{#1}
\providecommand{\url}[1]{\texttt{#1}}
\expandafter\ifx\csname urlstyle\endcsname\relax
  \providecommand{\doi}[1]{doi: #1}\else
  \providecommand{\doi}{doi: \begingroup \urlstyle{rm}\Url}\fi

\bibitem[Alon et~al.(2019)Alon, Livni, Malliaris, and Moran]{alon2019private}
Noga Alon, Roi Livni, Maryanthe Malliaris, and Shay Moran.
\newblock Private pac learning implies finite littlestone dimension.
\newblock In \emph{Proceedings of the 51st Annual ACM SIGACT Symposium on
  Theory of Computing}, pages 852--860, 2019.

\bibitem[Alon et~al.(2022)Alon, Bun, Livni, Malliaris, and
  Moran]{alon2022private}
Noga Alon, Mark Bun, Roi Livni, Maryanthe Malliaris, and Shay Moran.
\newblock Private and online learnability are equivalent.
\newblock \emph{ACM Journal of the ACM (JACM)}, 69\penalty0 (4):\penalty0
  1--34, 2022.

\bibitem[Baker(2016)]{baker20161}
Monya Baker.
\newblock 1,500 scientists lift the lid on reproducibility.
\newblock \emph{Nature}, 533\penalty0 (7604), 2016.

\bibitem[Ball(2023)]{ball2023ai}
Philip Ball.
\newblock Is ai leading to a reproducibility crisis in science?
\newblock \emph{Nature}, 624\penalty0 (7990):\penalty0 22--25, 2023.

\bibitem[Beimel et~al.(2013)Beimel, Nissim, and
  Stemmer]{beimel2013characterizing}
Amos Beimel, Kobbi Nissim, and Uri Stemmer.
\newblock Characterizing the sample complexity of private learners.
\newblock In Robert~D. Kleinberg, editor, \emph{Innovations in Theoretical
  Computer Science, {ITCS} '13, Berkeley, CA, USA, January 9-12, 2013}, pages
  97--110. {ACM}, 2013.
\newblock \doi{10.1145/2422436.2422450}.
\newblock URL \url{https://doi.org/10.1145/2422436.2422450}.

\bibitem[Blanc et~al.(2022{\natexlab{a}})Blanc, Lange, Malik, and
  Tan]{blanc2022popular}
Guy Blanc, Jane Lange, Ali Malik, and Li-Yang Tan.
\newblock Popular decision tree algorithms are provably noise tolerant.
\newblock In \emph{International Conference on Machine Learning}, pages
  2091--2106. PMLR, 2022{\natexlab{a}}.

\bibitem[Blanc et~al.(2022{\natexlab{b}})Blanc, Lange, Qiao, and
  Tan]{blanc2022properly}
Guy Blanc, Jane Lange, Mingda Qiao, and Li-Yang Tan.
\newblock Properly learning decision trees in almost polynomial time.
\newblock \emph{Journal of the ACM}, 69\penalty0 (6):\penalty0 1--19,
  2022{\natexlab{b}}.

\bibitem[Blanc et~al.(2023)Blanc, Lange, Malik, and Tan]{blanc2023lifting}
Guy Blanc, Jane Lange, Ali Malik, and Li-Yang Tan.
\newblock Lifting uniform learners via distributional decomposition.
\newblock In \emph{Proceedings of the 55th Annual ACM Symposium on Theory of
  Computing}, pages 1755--1767, 2023.

\bibitem[Blum et~al.(2003)Blum, Kalai, and Wasserman]{blum2003noise}
Avrim Blum, Adam Kalai, and Hal Wasserman.
\newblock Noise-tolerant learning, the parity problem, and the statistical
  query model.
\newblock \emph{Journal of the ACM (JACM)}, 50\penalty0 (4):\penalty0 506--519,
  2003.

\bibitem[Blum et~al.(2005)Blum, Dwork, McSherry, and Nissim]{blum2005practical}
Avrim Blum, Cynthia Dwork, Frank McSherry, and Kobbi Nissim.
\newblock Practical privacy: the sulq framework.
\newblock In \emph{Proceedings of the twenty-fourth ACM SIGMOD-SIGACT-SIGART
  symposium on Principles of database systems}, pages 128--138, 2005.

\bibitem[Blum(1994)]{blum1994separating}
Avrim~L Blum.
\newblock Separating distribution-free and mistake-bound learning models over
  the boolean domain.
\newblock \emph{SIAM Journal on Computing}, 23\penalty0 (5):\penalty0
  990--1000, 1994.

\bibitem[Bun(2020)]{bun2020computational}
Mark Bun.
\newblock A computational separation between private learning and online
  learning.
\newblock \emph{Advances in Neural Information Processing Systems},
  33:\penalty0 20732--20743, 2020.

\bibitem[Bun et~al.(2020)Bun, Livni, and Moran]{bun2020equivalence}
Mark Bun, Roi Livni, and Shay Moran.
\newblock An equivalence between private classification and online prediction.
\newblock In \emph{2020 IEEE 61st Annual Symposium on Foundations of Computer
  Science (FOCS)}, pages 389--402. IEEE, 2020.

\bibitem[Bun et~al.(2023)Bun, Gaboardi, Hopkins, Impagliazzo, Lei, Pitassi,
  Sivakumar, and Sorrell]{bun2023stability}
Mark Bun, Marco Gaboardi, Max Hopkins, Russell Impagliazzo, Rex Lei, Toniann
  Pitassi, Satchit Sivakumar, and Jessica Sorrell.
\newblock Stability is stable: Connections between replicability, privacy, and
  adaptive generalization.
\newblock In Barna Saha and Rocco~A. Servedio, editors, \emph{Proceedings of
  the 55th Annual {ACM} Symposium on Theory of Computing, {STOC} 2023, Orlando,
  FL, USA, June 20-23, 2023}, pages 520--527. {ACM}, 2023.
\newblock \doi{10.1145/3564246.3585246}.
\newblock URL \url{https://doi.org/10.1145/3564246.3585246}.

\bibitem[Bun et~al.(2024)Bun, Cohen, and Desai]{bun2024private}
Mark Bun, Aloni Cohen, and Rathin Desai.
\newblock Private {PAC} learning may be harder than online learning.
\newblock In Claire Vernade and Daniel Hsu, editors, \emph{International
  Conference on Algorithmic Learning Theory, 25-28 February 2024, La Jolla,
  California, {USA}}, volume 237 of \emph{Proceedings of Machine Learning
  Research}, pages 362--389. {PMLR}, 2024.
\newblock URL \url{https://proceedings.mlr.press/v237/bun24a.html}.

\bibitem[Canonne et~al.(2015)Canonne, Ron, and Servedio]{canonne2015testing}
Cl{\'e}ment~L Canonne, Dana Ron, and Rocco~A Servedio.
\newblock Testing probability distributions using conditional samples.
\newblock \emph{SIAM Journal on Computing}, 44\penalty0 (3):\penalty0 540--616,
  2015.

\bibitem[Canonne et~al.(2021)Canonne, Chen, Kamath, Levi, and
  Waingarten]{canonne2021random}
Cl{\'e}ment~L Canonne, Xi~Chen, Gautam Kamath, Amit Levi, and Erik Waingarten.
\newblock Random restrictions of high dimensional distributions and uniformity
  testing with subcube conditioning.
\newblock In \emph{Proceedings of the 2021 ACM-SIAM Symposium on Discrete
  Algorithms (SODA)}, pages 321--336. SIAM, 2021.

\bibitem[Chase et~al.(2023{\natexlab{a}})Chase, Chornomaz, Moran, and
  Yehudayoff]{chase2023local}
Zachary Chase, Bogdan Chornomaz, Shay Moran, and Amir Yehudayoff.
\newblock Local borsuk-ulam, stability, and replicability.
\newblock \emph{arXiv preprint arXiv:2311.01599}, 2023{\natexlab{a}}.

\bibitem[Chase et~al.(2023{\natexlab{b}})Chase, Moran, and
  Yehudayoff]{chase2023replicability}
Zachary Chase, Shay Moran, and Amir Yehudayoff.
\newblock Stability and replicability in learning.
\newblock In \emph{64th {IEEE} Annual Symposium on Foundations of Computer
  Science, {FOCS} 2023, Santa Cruz, CA, USA, November 6-9, 2023}, pages
  2430--2439. {IEEE}, 2023{\natexlab{b}}.
\newblock \doi{10.1109/FOCS57990.2023.00148}.
\newblock URL \url{https://doi.org/10.1109/FOCS57990.2023.00148}.

\bibitem[Dixon et~al.(2023)Dixon, Pavan, Woude, and
  Vinodchandran]{dixon2023list}
Peter Dixon, Aduri Pavan, Jason~Vander Woude, and N.~V. Vinodchandran.
\newblock List and certificate complexities in replicable learning.
\newblock In Alice Oh, Tristan Naumann, Amir Globerson, Kate Saenko, Moritz
  Hardt, and Sergey Levine, editors, \emph{Advances in Neural Information
  Processing Systems 36: Annual Conference on Neural Information Processing
  Systems 2023, NeurIPS 2023, New Orleans, LA, USA, December 10 - 16, 2023},
  2023.

\bibitem[Dvoretzky et~al.(1956)Dvoretzky, Kiefer, and
  Wolfowitz]{dvoretzky1956asymptotic}
Aryeh Dvoretzky, Jack Kiefer, and Jacob Wolfowitz.
\newblock Asymptotic minimax character of the sample distribution function and
  of the classical multinomial estimator.
\newblock \emph{The Annals of Mathematical Statistics}, pages 642--669, 1956.

\bibitem[Dwork et~al.(2006)Dwork, McSherry, Nissim, and
  Smith]{dwork2006calibrating}
Cynthia Dwork, Frank McSherry, Kobbi Nissim, and Adam Smith.
\newblock Calibrating noise to sensitivity in private data analysis.
\newblock In \emph{Theory of Cryptography: Third Theory of Cryptography
  Conference, TCC 2006, New York, NY, USA, March 4-7, 2006. Proceedings 3},
  pages 265--284. Springer, 2006.

\bibitem[Eaton et~al.(2023)Eaton, Hussing, Kearns, and
  Sorrell]{eaton2023replicable}
Eric Eaton, Marcel Hussing, Michael Kearns, and Jessica Sorrell.
\newblock Replicable reinforcement learning.
\newblock In Alice Oh, Tristan Naumann, Amir Globerson, Kate Saenko, Moritz
  Hardt, and Sergey Levine, editors, \emph{Advances in Neural Information
  Processing Systems 36: Annual Conference on Neural Information Processing
  Systems 2023, NeurIPS 2023, New Orleans, LA, USA, December 10 - 16, 2023},
  2023.

\bibitem[Esfandiari et~al.(2023{\natexlab{a}})Esfandiari, Kalavasis, Karbasi,
  Krause, Mirrokni, and Velegkas]{esfandiari2023replicableb}
Hossein Esfandiari, Alkis Kalavasis, Amin Karbasi, Andreas Krause, Vahab
  Mirrokni, and Grigoris Velegkas.
\newblock Replicable bandits.
\newblock In \emph{The Eleventh International Conference on Learning
  Representations}, 2023{\natexlab{a}}.

\bibitem[Esfandiari et~al.(2023{\natexlab{b}})Esfandiari, Karbasi, Mirrokni,
  Velegkas, and Zhou]{esfandiari2023replicable}
Hossein Esfandiari, Amin Karbasi, Vahab Mirrokni, Grigoris Velegkas, and Felix
  Zhou.
\newblock Replicable clustering.
\newblock In Alice Oh, Tristan Naumann, Amir Globerson, Kate Saenko, Moritz
  Hardt, and Sergey Levine, editors, \emph{Advances in Neural Information
  Processing Systems 36: Annual Conference on Neural Information Processing
  Systems 2023, NeurIPS 2023, New Orleans, LA, USA, December 10 - 16, 2023},
  2023{\natexlab{b}}.

\bibitem[Feldman et~al.(2017)Feldman, Grigorescu, Reyzin, Vempala, and
  Xiao]{feldman2017statistical}
Vitaly Feldman, Elena Grigorescu, Lev Reyzin, Santosh~S Vempala, and Ying Xiao.
\newblock Statistical algorithms and a lower bound for detecting planted
  cliques.
\newblock \emph{Journal of the ACM (JACM)}, 64\penalty0 (2):\penalty0 1--37,
  2017.

\bibitem[Fotakis et~al.(2020)Fotakis, Kalavasis, and
  Tzamos]{fotakis2020efficient}
Dimitris Fotakis, Alkis Kalavasis, and Christos Tzamos.
\newblock Efficient parameter estimation of truncated boolean product
  distributions.
\newblock In \emph{Conference on Learning Theory}, pages 1586--1600. PMLR,
  2020.

\bibitem[Fotakis et~al.(2021)Fotakis, Kalavasis, Kontonis, and
  Tzamos]{fotakis2021efficient}
Dimitris Fotakis, Alkis Kalavasis, Vasilis Kontonis, and Christos Tzamos.
\newblock Efficient algorithms for learning from coarse labels.
\newblock In \emph{Conference on Learning Theory}, pages 2060--2079. PMLR,
  2021.

\bibitem[Fotakis et~al.(2022)Fotakis, Kalavasis, and
  Tzamos]{fotakis2022perfect}
Dimitris Fotakis, Alkis Kalavasis, and Christos Tzamos.
\newblock Perfect sampling from pairwise comparisons.
\newblock In Sanmi Koyejo, S.~Mohamed, A.~Agarwal, Danielle Belgrave, K.~Cho,
  and A.~Oh, editors, \emph{Advances in Neural Information Processing Systems
  35: Annual Conference on Neural Information Processing Systems 2022, NeurIPS
  2022, New Orleans, LA, USA, November 28 - December 9, 2022}, 2022.

\bibitem[Georgiev and Hopkins(2022)]{georgiev2022privacy}
Kristian Georgiev and Samuel Hopkins.
\newblock Privacy induces robustness: Information-computation gaps and sparse
  mean estimation.
\newblock \emph{Advances in Neural Information Processing Systems},
  35:\penalty0 6829--6842, 2022.

\bibitem[Ghazi et~al.(2021{\natexlab{a}})Ghazi, Golowich, Kumar, and
  Manurangsi]{ghazi2021sample}
Badih Ghazi, Noah Golowich, Ravi Kumar, and Pasin Manurangsi.
\newblock Sample-efficient proper pac learning with approximate differential
  privacy.
\newblock In \emph{Proceedings of the 53rd Annual ACM SIGACT Symposium on
  Theory of Computing}, pages 183--196, 2021{\natexlab{a}}.

\bibitem[Ghazi et~al.(2021{\natexlab{b}})Ghazi, Kumar, and
  Manurangsi]{ghazi2021user}
Badih Ghazi, Ravi Kumar, and Pasin Manurangsi.
\newblock User-level differentially private learning via correlated sampling.
\newblock In Marc'Aurelio Ranzato, Alina Beygelzimer, Yann~N. Dauphin, Percy
  Liang, and Jennifer~Wortman Vaughan, editors, \emph{Advances in Neural
  Information Processing Systems 34: Annual Conference on Neural Information
  Processing Systems 2021, NeurIPS 2021, December 6-14, 2021, virtual}, pages
  20172--20184, 2021{\natexlab{b}}.

\bibitem[Goel et~al.(2020)Goel, Gollakota, and Klivans]{goel2020statistical}
Surbhi Goel, Aravind Gollakota, and Adam Klivans.
\newblock Statistical-query lower bounds via functional gradients.
\newblock \emph{Advances in Neural Information Processing Systems},
  33:\penalty0 2147--2158, 2020.

\bibitem[Goldreich et~al.(1986)Goldreich, Goldwasser, and
  Micali]{goldreich1986construct}
Oded Goldreich, Shafi Goldwasser, and Silvio Micali.
\newblock How to construct random functions.
\newblock \emph{Journal of the ACM (JACM)}, 33\penalty0 (4):\penalty0 792--807,
  1986.

\bibitem[Gonen et~al.(2019)Gonen, Hazan, and Moran]{gonen2019private}
Alon Gonen, Elad Hazan, and Shay Moran.
\newblock Private learning implies online learning: An efficient reduction.
\newblock \emph{Advances in Neural Information Processing Systems}, 32, 2019.

\bibitem[Gouleakis et~al.(2017)Gouleakis, Tzamos, and
  Zampetakis]{gouleakis2017faster}
Themistoklis Gouleakis, Christos Tzamos, and Manolis Zampetakis.
\newblock Faster sublinear algorithms using conditional sampling.
\newblock In \emph{Proceedings of the Twenty-Eighth Annual ACM-SIAM Symposium
  on Discrete Algorithms}, pages 1743--1757. SIAM, 2017.

\bibitem[Gupta et~al.(2011)Gupta, Hardt, Roth, and Ullman]{gupta2011privately}
Anupam Gupta, Moritz Hardt, Aaron Roth, and Jonathan Ullman.
\newblock Privately releasing conjunctions and the statistical query barrier.
\newblock In \emph{Proceedings of the forty-third annual ACM symposium on
  Theory of computing}, pages 803--812, 2011.

\bibitem[Impagliazzo et~al.(2022)Impagliazzo, Lei, Pitassi, and
  Sorrell]{impagliazzo2022reproducibility}
Russell Impagliazzo, Rex Lei, Toniann Pitassi, and Jessica Sorrell.
\newblock Reproducibility in learning.
\newblock In Stefano Leonardi and Anupam Gupta, editors, \emph{{STOC} '22: 54th
  Annual {ACM} {SIGACT} Symposium on Theory of Computing, Rome, Italy, June 20
  - 24, 2022}, pages 818--831. {ACM}, 2022.
\newblock \doi{10.1145/3519935.3519973}.
\newblock URL \url{https://doi.org/10.1145/3519935.3519973}.

\bibitem[Janson(2018)]{janson2018tail}
Svante Janson.
\newblock Tail bounds for sums of geometric and exponential variables.
\newblock \emph{Statistics \& Probability Letters}, 135:\penalty0 1--6, 2018.

\bibitem[Kalavasis et~al.(2023)Kalavasis, Karbasi, Moran, and
  Velegkas]{kalavasis2023statistical}
Alkis Kalavasis, Amin Karbasi, Shay Moran, and Grigoris Velegkas.
\newblock Statistical indistinguishability of learning algorithms.
\newblock In Andreas Krause, Emma Brunskill, Kyunghyun Cho, Barbara Engelhardt,
  Sivan Sabato, and Jonathan Scarlett, editors, \emph{International Conference
  on Machine Learning, {ICML} 2023, 23-29 July 2023, Honolulu, Hawaii, {USA}},
  volume 202 of \emph{Proceedings of Machine Learning Research}, pages
  15586--15622. {PMLR}, 2023.
\newblock URL \url{https://proceedings.mlr.press/v202/kalavasis23a.html}.

\bibitem[Kalavasis et~al.(2024)Kalavasis, Karbasi, Larsen, Velegkas, and
  Zhou]{kalavasis2024replicable}
Alkis Kalavasis, Amin Karbasi, Kasper~Green Larsen, Grigoris Velegkas, and
  Felix Zhou.
\newblock Replicable learning of large-margin halfspaces.
\newblock \emph{arXiv preprint arXiv:2402.13857}, 2024.

\bibitem[Karbasi et~al.(2023)Karbasi, Velegkas, Yang, and
  Zhou]{karbasi2023replicability}
Amin Karbasi, Grigoris Velegkas, Lin Yang, and Felix Zhou.
\newblock Replicability in reinforcement learning.
\newblock In Alice Oh, Tristan Naumann, Amir Globerson, Kate Saenko, Moritz
  Hardt, and Sergey Levine, editors, \emph{Advances in Neural Information
  Processing Systems 36: Annual Conference on Neural Information Processing
  Systems 2023, NeurIPS 2023, New Orleans, LA, USA, December 10 - 16, 2023},
  2023.

\bibitem[Kasiviswanathan et~al.(2011)Kasiviswanathan, Lee, Nissim,
  Raskhodnikova, and Smith]{kasiviswanathan2011can}
Shiva~Prasad Kasiviswanathan, Homin~K Lee, Kobbi Nissim, Sofya Raskhodnikova,
  and Adam Smith.
\newblock What can we learn privately?
\newblock \emph{SIAM Journal on Computing}, 40\penalty0 (3):\penalty0 793--826,
  2011.

\bibitem[Kearns(1998)]{kearns1998efficient}
Michael Kearns.
\newblock Efficient noise-tolerant learning from statistical queries.
\newblock \emph{Journal of the ACM (JACM)}, 45\penalty0 (6):\penalty0
  983--1006, 1998.

\bibitem[Kearns and Vazirani(1994)]{kearns1994introduction}
Michael~J Kearns and Umesh Vazirani.
\newblock \emph{An introduction to computational learning theory}.
\newblock MIT press, 1994.

\bibitem[Littlestone(1988)]{littlestone1988learning}
Nick Littlestone.
\newblock Learning quickly when irrelevant attributes abound: A new
  linear-threshold algorithm.
\newblock \emph{Machine learning}, 2:\penalty0 285--318, 1988.

\bibitem[Massart(1990)]{massart1990tight}
Pascal Massart.
\newblock The tight constant in the dvoretzky-kiefer-wolfowitz inequality.
\newblock \emph{The annals of Probability}, pages 1269--1283, 1990.

\bibitem[Moran et~al.(2023)Moran, Schefler, and Shafer]{moran2023bayesian}
Shay Moran, Hilla Schefler, and Jonathan Shafer.
\newblock The bayesian stability zoo.
\newblock In Alice Oh, Tristan Naumann, Amir Globerson, Kate Saenko, Moritz
  Hardt, and Sergey Levine, editors, \emph{Advances in Neural Information
  Processing Systems 36: Annual Conference on Neural Information Processing
  Systems 2023, NeurIPS 2023, New Orleans, LA, USA, December 10 - 16, 2023},
  2023.

\bibitem[Pineau et~al.(2019)Pineau, Sinha, Fried, Ke, and
  Larochelle]{pineau2019iclr}
Joelle Pineau, Koustuv Sinha, Genevieve Fried, Rosemary~Nan Ke, and Hugo
  Larochelle.
\newblock Iclr reproducibility challenge 2019.
\newblock \emph{ReScience C}, 5\penalty0 (2):\penalty0 5, 2019.

\end{thebibliography}
\endgroup

\newpage
\appendix

\section{Notation and Preliminaries}

For standard PAC learning definitions, we refer to the book of \citet{kearns1994introduction}.

\subsection{Replicable Learning}
Following the pioneering work of \citet{impagliazzo2022reproducibility},
we consider the definition of a replicable learning algorithm.
\replicableDefinition*

In words,
$\mcal A$ is replicable if sharing the randomness across two executions on different i.i.d. datasets
yields the exact same output with high probability.
We can think of $\mcal A$ as a training algorithm which is replicable
if by fixing the random seed,
it outputs the exact same model with high probability.

One of the most elementary statistical operations
we may wish to make replicable is mean estimation.
This operation can be phrased more broadly using the language of \emph{statistical queries}.
\begin{defn}[Statistical Query Oracle; \citealp{kearns1998efficient}]\label{def:statistical query oracle} 
    Let $\mathcal{D}$ be a distribution over the domain $\mathcal{X}$
    and $\phi: \mathcal{X} \to \R$ be a statistical query
    with true value
    $
        v^\star := \lim_{n\to \infty} \phi(X_1, \dots, X_n)\in \R.
    $
    Here $X_i\sim_{i.i.d.} \mcal D$
    and the convergence is understood in probability or distribution.
    Let $\alpha,\beta \in (0,1)^2$.
    A \emph{statistical query (SQ) oracle} outputs 
    a value $v$ such that $\abs*{v - v^\star} \leq \alpha$
    with probability at least $1-\beta$.
\end{defn}

The SQ framework appears in various learning theory contexts (see e.g., \citet{blum2003noise,gupta2011privately,goel2020statistical,fotakis2021efficient} and the references therein). In the SQ model, the learner interacts with an oracle
in the following way: the learner submits a statistical query to the oracle
and the oracle returns its true value, after adding some noise to it. 

The simplest example of a statistical query is the sample mean
$
    \phi(X_1, \dots, X_n)
    = \frac1n \sum_{i=1}^n X_i.
$
\citet{impagliazzo2022reproducibility} designed a replicable SQ-query oracle
for sample mean queries with bounded co-domain.
\citet{esfandiari2023replicable} generalized the result to simultaneously estimate the means of multiple random variables with unbounded co-domain 
under some regularity conditions on their distributions (cf. \Cref{thm:replicable rounding}).
The idea behind both results is to use a replicable rounding technique introduced in \citet{impagliazzo2022reproducibility} which allows one to sacrifice some accuracy
of the estimator in exchange for the replicability property.

\subsection{Private Learning}
\paragraph{Differential Privacy.}
A foundational notion of algorithmic
stability is that of Differential Privacy (DP)
\citep{dwork2006calibrating}.
For $a,b, \varepsilon,\delta \in (0,1)$,
let $a \approx_{\varepsilon, \delta} b$ denote the statement $a \leq e^\varepsilon b + \delta$ and $b \leq e^\varepsilon a + \delta$. We say that two probability distributions $P, Q$ are $(\varepsilon, \delta)$-indistinguishable if $P(E) \approx_{\varepsilon, \delta} Q(E)$ for any measurable event $E$. 
\begin{defn}
[Approximate DP Algorithm; \citealp{kasiviswanathan2011can}]
\label{def:dp}
A learning algorithm $A$
is an $n$-sample $(\varepsilon,\delta)$-differentially private if for any pair of samples $S, S' \in (\mcal X \times \{0,1\})^n$ that disagree on a single example, the induced posterior distributions $A(S)$ and $A(S')$ are $(\varepsilon, \delta)$-indistinguishable.
\end{defn}

In the previous definition, when the 
parameter $\delta = 0,$ we say that the
algorithm satisfies (\emph{pure}) $\varepsilon$-DP.

We remind the reader that, in the context of PAC learning, any hypothesis class $\mscr C$ can be PAC-learned by an approximate differentially private algorithm if and only if it has finite \emph{Littlestone dimension}, i.e., there is a qualitative equivalence between online learnability and private PAC learnability \citep{alon2019private,bun2020equivalence,ghazi2021sample,alon2022private}.
\subsection{Online Learning}\label{sec:online preliminary}
We consider the no-regret model of online learning.
Recall Littlestone's model of (realizable) online learning \citep{littlestone1988learning}
defined via a two-player game between a learner and an adversary.
Let $\mscr C$ be a given concept class.

In each round of the game $t=1, \dots, T$
where $T$ is a time horizon known to a (randomized) learner, the interaction
is the following:
\begin{enumerate}[1)]
  \item The adversary selects some features $x_t\in \Set{0, 1}^d$.
  \item The learner predicts a label $\hat b_t\in \Set{0, 1}$, potentially
  using randomization,
  \item The adversary, who observes the distribution of the choice of the learner
  but not its realization, chooses the correct label $b_t = c(x_t)$ under the constraint
  that there exists some $c^\star_t \in \mscr C$ such that $c^\star_\tau(x_\tau) = b_\tau$, for all $\tau \leq t$.
\end{enumerate}
The goal of the learner is to minimize its regret,
defined by
\begin{align*}
  R_T
  &:= \max_{h\in \mscr C} \left\{ \sum_{t=1}^T \ones\Set{\hat b_t\neq c(x_t)} -  \ones\Set{h(x_t)\neq c(x_t)} \right\} \\
  &= \sum_{t=1}^T \ones\Set{\hat b_t\neq c(x_t)} - 0 &&h=c \\
  &= \sum_{t=1}^T \ones\Set{\hat b_t\neq c(x_t)}.
\end{align*}
Thus the learner aims to compete with the best concept in hindsight from the class $\mscr C$.
However,
realizability ensures that the best such concept makes no mistakes
so the regret is defined only in terms of the number of mistakes of the learner.

We now introduce the definition of an \emph{efficient no-regret} learning algorithm used by \citet{bun2020computational}.
In the definition below,
we write $\card{c}$ to denote the length of a minimal description of the concept $c$.
\begin{defn}[Efficient No-Regret Learning; \citep{bun2020computational}]
We say that a learner \emph{efficiently no-regret learns $\mscr C$}
  if there is some $\eta > 0$
  such that for every adversary,
  it achieves expected regret
  \[
    \E[R_T]
    = \poly(d, \card{c}) T^{1-\eta}
  \]
  using time $\poly(d, \card{c}, T)$ in every round.
\end{defn}
There are two non-standard features of this definition.
First,
no-regret algorithms are typically only required to achieve sublinear regret $o(T)$ in $T$,
whereas we require it to be strongly sublinear $T^{1-\eta}$.
A stronger condition like this is needed to make the definition nontrivial since the sample space is finite.
Indeed,
suppose $T = 2^d$,
then the trivial random guessing algorithm attains a regret bound of
\begin{align*}
  \frac{T}2
  &= \frac{dT}{2d} \\
  &= \frac{d T}{2\log T} \\
  &= \poly(d)o(T).
\end{align*}
Many no-regret algorithms such as the multiplicative weights update algorithm achieve strongly sublinear regret.

Second,
it would be more natural to require the learner to run in $\poly(\log T)$ time,
the description length of the time horizon,
rather than the value of $T$ itself.
The relaxed formulation only makes the separation stronger.

\section{Efficient Replicability and Online Learning}
\label{sec:replicable but not online}

\subsection{One-Way Sequences}
Our exposition follows that of~\citet{bun2020computational}.
For every dimension $d \in \N$,
\citet{blum1994separating} defines a concept class $\OWS_d$
consisting of functions over the domain $\Set{0, 1}^d$
that can be represented using $\poly(d)$ bits and evaluated in $\poly(d)$ time.
The concepts of $\OWS_d$ are indexed by bit strings $s\in \Set{0, 1}^k$,
where $k = \floor{\sqrt{d}}-1$
\[
  \OWS_d = \Set*{c_s: s\in \Set{0, 1}^k}.
\]
We will usually omit the index $d$ when it is clear from the context.
Each $c_s: \Set{0, 1}^d\to \Set{0, 1}$
is defined using two efficiently representable and computable functions
\begin{align*}
  G &: \Set{0, 1}^k\times \Set{0, 1}^k\to \Set{0, 1}^{d-k}\,, \\
  f &: \Set{0, 1}^k\times \Set{0, 1}^k\to \Set{0, 1}
\end{align*}
 that are based on the Goldreich-Goldwasser-Micali pseudorandom function family \citep{goldreich1986construct}.
We omit the definition of these functions since it does not impede us towards our goal
and refer the reader to \citet{blum1994separating} for details.
Intuitively,
$G(i, s)$ computes the string $\sigma_i$ described in the introduction
and $f(i, s)$ computes its label $b_i$.
We can think of $s$ as the random seed which is generated from some source of true randomness
which is then used to construct $G, f$.

For convenience,
we identify $\Set{0, 1}^k \equiv [2^k]$.
Then $c_s$ is defined as
\[
  c_s(i, \sigma) =
  \begin{cases}
    1, &G(i, s) = \sigma, f(i, s) = 1\,, \\
    0, &\text{else}\,.
  \end{cases}
\]
We see that $c_s(i, \sigma)$ encodes both the string $\sigma_i$
as well as its label $b_i$,
for every random seed $s\in \Set{0, 1}^k$.

The two relevant properties of the strings $\sigma_i$ are summarized below.
\begin{prop}[Forward is Easy; \citealp{blum1994separating}]\label{prop:compute forward}
  There is an efficiently computable function
  \[
    \ComputeForward: \Set{0, 1}^k\times \Set{0, 1}^k\times \Set{0, 1}^{d-k} \to \Set{0, 1}^{d-k}\times \Set{0, 1}
  \]
  such that for every $j > i$,
  \[
    \ComputeForward(j, i, G(i, s)) = (G(j, s), f(j, s)).
  \]
\end{prop}

\begin{prop}[Reverse is Hard; \citealp{blum1994separating}]\label{prop:no backward}
  Assuming the existence of one-way functions,
  there exist functions $G: \Set{0, 1}^k\times \Set{0, 1}^k\to \Set{0, 1}^{d-k}$
  and $f: \Set{0, 1}^k\times \Set{0, 1}^k\to \Set{0, 1}$
  satisfying the following.
  Let $\mcal O$ be an oracle that on input $(j, i, G(i, s))$,
  outputs $(G(j, s), f(j, s))$
  for any $j > i$.
  Let $A$ be any polynomial time randomized algorithm
  and let $A^{\mcal O}$ denote the algorithm with access to the oracle $\mcal O$.
  For every $i\in \Set{0, 1}^k$,
  \[
    \Pr\left[ A^{\mcal O}(i, G(i, s)) = f(i, s) \right]
    \leq \frac12 + \negl(d),
  \]
  where the probability is taken over the internal randomness of $A$
  and uniformly random $s\sim \mcal U\left(\Set{0, 1}^k\right)$.
\end{prop}
The above proposition states that no efficient algorithm can outperform random guessing when trying determinew the label of a string,
even given access to a ``compute-forward'' oracle.

\subsection{Hardness of Efficiently Online Learning \texorpdfstring{$\OWS$}{OWSd}}
\citet{blum1994separating} used \Cref{prop:no backward} to show that \OWS cannot be learned in the \emph{mistake bound} model,
a more stringent model of online learning compared to the no-regret setting.
Later, \citet{bun2020computational} adapted the argument to the no-regret setting,
making the separation stronger.
\begin{thm}[\citep{bun2020computational}]\label{thm:online hardness OWS}
  Assuming the existence of one-way functions,
  \OWS cannot be learned by an efficient no-regret algorithm.
\end{thm}

Let $\card{c}$ denote the length of a minimal description of the concept $c$.
Recall that 
a learner \emph{efficiently no-regret learns a class $\mathscr C$} if there exists $\eta > 0$ such that for every adversary and any $c \in \mathscr C$,
it achieves 
$\mathbb E[R_T] = \poly(d, |c|) \cdot T^{1-\eta}$ (strongly sublinear) using time $\poly(d, |c|, T)$ in every round.
As mentioned in \Cref{sec:online preliminary},
this requirement of strongly sublinear regret is necessary to make the definition non-trivial
as the random guessing algorithm attains $o(T)$ regret in finite domains.

\subsection{Replicable Query Rounding}
\citet{impagliazzo2022reproducibility} designed replicable statistical query oracles (cf. \Cref{def:statistical query oracle}) for bounded co-domains
and \citet{esfandiari2023replicable} generalized their results to multiple general queries with unbounded co-domain (cf. \Cref{thm:replicable rounding}),
assuming some regularity conditions on the queries.

We illustrate the idea behind the rounding procedure applied to the task of mean estimation.
An initial observation is that across two executions,
the empirical mean concentrates about the true mean with high probability.
Next,
we discretize the real line into equal-length intervals
with a random shift.
It can be shown that both points fall into the same random interval with high probability
and outputting the midpoint of said interval yields the replicability guarantee.

For completeness,
we include the pseudocode and a proof of the replicable rounding procedure.
\begin{restatable}[Replicable Rounding; \citealp{impagliazzo2022reproducibility, esfandiari2023replicable}]{thm}{rRounding}\label{thm:replicable rounding}
  Let $\mcal D$ be a distribution over some domain $\mcal X$.
  Let $\alpha, \rho\in (0, 1)$ and $\beta\in (0, \nicefrac\rho3)$.
  Suppose we have a sequence of statistical queries $g_1, \dots, g_T: \mcal X\to \R$
  with true values $\mu_1, \dots, \mu_T$
  and sampling $n$ independent points from $\mcal D$
  ensures that
  \[
    \max_{t\in [T]} \abs{g_t(x_1, \dots, x_n) - \mu_t} \leq \alpha
  \]
  with probability at least $1-\beta$.

  \sloppy Then there is a polynomial-time time postprocessing procedure \rRound (cf. \Cref{alg:replicable SQ})
  such that each composition $\rRound(g_t, \alpha, \rho)$ is $\rho$-replicable.
  Moreover,
  the outputs $\hat \mu_t = \rRound(g_t(x_1, \dots, x_n), \alpha, \rho)$ satisfy
  \[
    \max_{t\in [T]} \abs{\wh \mu_t - \mu_t}\leq \frac{4\alpha}\rho
  \]
  with probability at least $1-\beta$.
\end{restatable}
Note that if we wish for the sequence of rounding steps to be $\rho$-replicable overall,
we can simply run each rounding step with parameter $\nicefrac{\rho}T$.

\begin{algorithm}[H]
\caption{Replicable Rounding}\label{alg:replicable SQ}
\begin{algorithmic}[1]
  \STATE {\rRound}{(query values $g_1, \dots, g_T$, accuracy $\alpha$, replicability $\rho$):}
  \STATE $L\gets \nicefrac{6\alpha}\rho$
  \STATE Sample $L_0\sim U[0, L)$
  \STATE Discretize the real line $\dots, [L_0-L, L_0), [L_0, L_0+L), [L_0+L, L_0+2L), \dots$ into disjoint intervals
  \FOR {$t=1, \dots, T$}
    \STATE Round $g_t$ to the midpoint of the interval, $\hat \mu_t$
  \ENDFOR
  \RETURN $\hat \mu\in \R^T$.
\end{algorithmic}
\end{algorithm}

\begin{pf}[\Cref{thm:replicable rounding}]
  Discretize the real line as disjoint intervals.
  \[
    \dots, [-L, 0), [0, L), [L, 2L), \dots.
  \]
  We will choose the value of $L$ later. 
  Consider adding a uniformly random offset $L_0\sim U[0, L)$ so the discretization becomes
  \[
    \dots, [L_0-L, L_0), [L_0, L_0+L), [L_0+L, L_0+2L), \dots.
  \]
  For each $t\in T$,
  we round the estimate $g_t(x_{1:n}) := g_t(x_1, \dots, x_n)$ to the midpoint of the interval it falls into.
  Let $\wh \mu_t$ be the rounded estimate that we output.
  From hereonforth,
  we condition on the event $\max_t \abs{g_t(x_{1:n}) - \mu_t} \leq \alpha$.

  \paragraph{Replicability of \Cref{alg:replicable SQ}.}
  Fix $t\in T$.
  Consider the output across two runs $\hat \mu_t, \hat \mu_t'$.
  As long as the raw estimates $g_t(x_{1:n}), g_t(x_{1:n}')$ fall in the same interval,
  the outputs will be exactly the same.
  This occurs with probability
  \[
    \frac{\abs{g_t(x_{1:n}) - g_t(x_{1:n}')}}{L}
    \leq \frac{2\alpha}{L}.
  \]
  Choosing $L = \nicefrac{6\alpha}\rho$ ensures this value is at most $\nicefrac\rho3$.
  Accounting for the $2\beta\leq \nicefrac{2\rho}3$ probability of the estimates $g_t(x_{1:n})$
  failing to concentrate,
  the output $\wh \mu_t$ is $\rho$-replicable.

  \paragraph{Correctness of \Cref{alg:replicable SQ}.}
  The rounding incurs an additive error of at most $\nicefrac{L}2$.
  The choice of $L = \nicefrac{6\alpha}{\rho}$ means the total error is at most
  \[
    \alpha + \frac{3\alpha}{\rho}
    \leq \frac{4\alpha}\rho.
  \]
\end{pf}

\subsection{Replicable Quantile Estimation}
In this section we provide an algorithm that replicably estimates \emph{quantiles}
of a distribution. This will be useful in providing the computational
separation between replicable PAC learning and online learning. 
We believe that it can have applications beyond the scope of our work.

We first present a well-known concentration inequality regarding CDFs of random variables
due to Dvoretzky, Kiefer, and Wolfowitz. 

\begin{thm}[Dvoretzky–Kiefer–Wolfowitz Inequality; \citealp{dvoretzky1956asymptotic, massart1990tight}]\label{thm:DKW ineq}
  Let $X_1, \dots, X_n$ be i.i.d. random variables
  with CDF $F$.
  Let $F_n$ denote the empirical distribution function given by
  $
    F_n(x)
    := \frac1n \sum_{i=1}^n \ones\Set{X_i\leq x}.
  $
  For any $\alpha > 0$,
  \[
    \Pr\left[ \sup_{x\in \R} \abs{F_n(x) - F(x)} > \alpha \right]
    \leq 2\exp(-2n\alpha^2).
  \]
\end{thm}
In other words,
we require at most
$
  n = (\nicefrac1{2\alpha^2})\ln(\nicefrac2\beta)
$
samples to ensure that with probability at least $1-\beta$,
the empirical CDF uniformly estimates the true CDF
with error at most $\alpha$.

The replicable quantile estimation algorithm for discrete
and bounded distributions can be found
in \Cref{alg:replicable quantile est}. The high-level
idea is to perform a (replicable) binary search 
over the support of the distribution and to check
whether the empirical CDF evaluated at some point $x$ is above or below
the target quantile $q.$
\begin{algorithm}[H]
\caption{Replicable Quantile Estimation}\label{alg:replicable quantile est}
\begin{algorithmic}[1]
  \STATE {\rQuantileEst}{(samples $x_1, \dots, x_n$, quantile $q$, accuracy $\alpha$, replicability $\rho$, confidence $\beta$):}
  \STATE $F_n(i) \gets \sum_{j=1}^n \ones\Set{x_j\leq i}$ \COMMENT{Implicitly for all $i\in [R]$}
  \STATE $\ell \gets 0$; $h \gets R$
  \WHILE {$\ell < h-1$}
    \STATE $m \gets \nicefrac{(\ell+h)}2$
    \STATE $\widetilde F_n(m) \gets \rRound(F_n(m), \nicefrac{\alpha\rho}{4\log_2(R)}, \nicefrac{\rho}{\log_2(R)})$
    \IF {$\widetilde F_n(m) \geq q$}
      \STATE $h \gets m$
    \ELSE
      \STATE $\ell \gets m$
    \ENDIF
  \ENDWHILE
  \RETURN $h$.
\end{algorithmic}
\end{algorithm}

\begin{thm}[Replicable Quantile Estimation]\label{thm:replicable quantile est}
  Let $\alpha, \rho\in (0, 1)$ and $\beta\in (0, \nicefrac{\rho}3)$.
  Suppose we have access to
  \[
    m = \frac{16 \log_2^2(R)}{2\alpha^2\rho^2} \ln\frac2\beta
    = O\left( \frac{\log^2 R}{\alpha^2 \rho^2}{\ln\frac1\beta} \right)
  \]
  i.i.d. samples from some distribution over $[R]$ with CDF $F$
  and $q\in [0, 1]$ is the desired quantile level.
  \Cref{alg:replicable quantile est} terminates in $O(\poly(n))$ time,
  is $\rho$-replicable,
  and with probability at least $1-\beta$,
  outputs some $x\in [R]$ such that
  \begin{align*}
    F(x) \geq q - \alpha, \qquad
    F(x-1) < q + \alpha.
  \end{align*}
\end{thm}
Our replicable quantile estimation differs from the replicable median algorithm of \citep{impagliazzo2022reproducibility} in at least two ways. 
Firstly, the replicable median algorithm of \citep{impagliazzo2022reproducibility} seems to rely heavily on properties of (approximate) medians in order to satisfy the approximation guarantees. 
On the other hand, our algorithm works regardless of the desired quantile. 
Secondly, their median algorithm relies on a non-trivial recursive procedure while our replicable quantile algorithm is considerably simpler and is based on a concentration of the CDF through the DKW inequality.

\begin{pf}[\Cref{thm:replicable quantile est}]
  By the DKW inequality (\Cref{thm:DKW ineq}),
  sampling
  \[
    m = \frac1{2(\nicefrac{\alpha\rho}{4\log_2(R)})^2} \ln\frac2\beta
  \]
  points from our distribution ensures that 
  \[
    \sup_{x\in [R]} \abs{F_n(x) - F(x)} \leq \frac{\alpha\rho}{4\log_2(R)}
  \]
  with probability at least $1-\beta$. The proof
  is divided into two steps. We first show
  the correctness of our algorithm and then argue
  about its replicability.
  \paragraph{Correctness of \Cref{alg:replicable quantile est}.}
  The loop invariant we wish to maintain is that
  \begin{align*}
    F(h) \geq q - \alpha, \qquad
    F(\ell) < q + \alpha.
  \end{align*}
  This is certainly initially true since our distribution is over $[R] = \Set{1, \dots, R}$
  and hence there is no mass at or below $\ell = 0$
  while all the mass is at or below $h = R$.

  Let $m_1, \dots, m_{T}$ denote the midpoints chosen by binary search
  with $T = \log_2(R)$.
  By \Cref{thm:replicable rounding},
  if we condition on the success of all executions of $\rRound(F_n(m_t), \nicefrac{\alpha\rho}{4\log_2(R)}, \nicefrac{\rho}{\log_2(R)})$,
  then their outputs $\widetilde F_n(m_t)$ estimate $F(m_t)$ with additive error at most
  \[
    \frac{4\nicefrac{\alpha\rho}{4\log_2(R)}}{\nicefrac{\rho}{\log_2(R)}}
    = \alpha.
  \]
  We update $h\gets m_t$ only if
  \[
    q \leq \widetilde F_n(m_t) \leq F(m_t) + \alpha
    \implies q-\alpha \leq F(m_t).
  \]
  Similarly,
  we update $\ell \gets m_t$ only if
  \[
    q > \widetilde F_n(m_t) \geq F(m_t) - \alpha
    \implies q+\alpha > F(m_t).
  \]
  Hence the loop invariant is maintained.

  At termination,
  $\ell = h - 1$ with
  \begin{align*}
    F(h) &\geq q - \alpha \\
    F(h-1) &< q + \alpha
  \end{align*}
  by the loop invariant as desired.

  \paragraph{Replicability of \Cref{alg:replicable quantile est}.}
  By \Cref{thm:replicable rounding},
  assuming all prior branching decisions are identical,
  each new branching decision in \Cref{alg:replicable quantile est} is $(\nicefrac{\rho}{\log_2(R)})$-replicable.
  Since we make at most $\log_2(R)$ decisions,
  the entire algorithm is $\rho$-replicable.
\end{pf}

\subsection{An Efficient Replicable Learner for \texorpdfstring{$\OWS_d$}{OWSd}}
Equipped with the replicable quantile estimator from \Cref{thm:replicable quantile est},
we are ready to present our efficient replicable $\OWS_d$ PAC learner. We outline
below a high-level description of our algorithm which can be found in \Cref{alg:replicable learner OWS}.
\begin{enumerate}[1),itemsep=0em]
  \item We first replicably estimate the mass of all the elements that have positive labels.
    If it is much smaller than some threshold $O(\alpha)$, then we output the zero hypothesis.
  \item Then we take sufficiently many samples to get enough data points with positive label
    and we run the replicable quantile estimation on the marginal distribution of features with positive label to get some $x\in [2^k]$.
  \item Next, we take sufficiently many samples to get a positive point at or below $x$
    and then we forward compute the label of $x$.
  \item The hypothesis we return is the following: If its input $(i, \sigma)$ has index less than $i^*$, it outputs 0.
    If its input is exactly $(i^*, \sigma^*)$, it outputs the previously computed label.
    Otherwise, it forward computes the label using $i, \sigma$.
\end{enumerate}

\begin{algorithm}[H]
\caption{Replicable Learner for $\OWS_d$}\label{alg:replicable learner OWS}
\begin{algorithmic}[1]
  \STATE {\rLearnerOWSd}{(samples $(i_1, \sigma_1), \dots, (i_n, \sigma_m)$, accuracy $\alpha$, replicability $\rho$, confidence $\beta$):}
  \STATE Let $S_+ := (i_{j_1}, \sigma_{j_1}), \dots, (i_{j_n}, \sigma_{j_n})$ be the subsequence of positive samples,
  where $i_{j_k} \leq i_{j_{k+1}}$.
  \STATE $\wh p \gets \rRound(\nicefrac{n}{m}, \nicefrac{\rho\alpha}{48}, \nicefrac\rho3)$
  \IF {$\wh p < \nicefrac\alpha2$} \label{state:positive threshold}
    \RETURN All-zero hypothesis. \label{state:return all-zero}
  \ENDIF
  \STATE $i^* \gets \rQuantileEst(\Set{i_{j_1}, \dots, i_{j_n}}, \nicefrac\alpha2, \nicefrac\alpha4, \nicefrac\rho3, \nicefrac\beta3)$
  \IF {$i_{j_1} \geq i^*$}
    \RETURN ``FAILURE''
  \ENDIF
  \STATE $(\sigma^*, b^*) \gets \ComputeForward(i^*, i_{j_1}, \sigma_{j_1})$
  \RETURN hypothesis $h(i, \sigma) :=$ `` \\
  If $i < i^*$, output $0$. \\
  If $i=i^*$, output $b^*$ if $\sigma^* = \sigma$ and $0$ otherwise. \\
  If $i > i^*$, get $(\hat\sigma, \hat b) \gets \ComputeForward(i, i^*, \sigma^*)$
  and output $\hat b$ if $\sigma = \hat\sigma$ and $0$ otherwise.''
\end{algorithmic}
\end{algorithm}

\begin{thm}\label{thm:replicable learner OWS}
  Let $\rho, \alpha\in (0, 1)$ and $\beta\in (0, \nicefrac{\rho}3)$.
  \Cref{alg:replicable learner OWS} is an efficient $\rho$-replicable $(\alpha, \beta)$-PAC learner for $\OWS_d$
  with sample complexity
  \[
    m
    = \max\left( \frac{392}{\alpha^2 \rho^2}\ln\frac6\beta,
    \frac{9216 k^2}{\alpha^3 \rho^2} \ln\frac6\beta,
    \frac{32}{\alpha^2}\ln\frac6\beta \right)
    = O\left( \frac{d^2}{\alpha^3 \rho^2}\ln\frac1\beta \right)
  \]
  and time complexity
  \[
    O(\poly(m)).
  \]
\end{thm}

\begin{pf}[\Cref{thm:replicable learner OWS}]
  The indicator random variable $I_i := \ones\Set{\sigma_i = 1}$ is a bounded random variable
  and its sample mean is precisely $\nicefrac{n}{m}$.
  Let $p\in [0, 1]$ denote the probability mass of the positively labeled elements.
  By an Hoeffding inequality,
  $\abs{\nicefrac{n}{m} - p}\leq \alpha$ with probability at least 
  \[
    2 \exp\left( -2m\alpha^2 \right).
  \]
  Since we have
  \[
    m
    \geq \frac{392}{\alpha^2 \rho^2} \ln\frac6\beta,
  \]
  then $\abs{\nicefrac{m}{n} - p}\leq \nicefrac{\rho\alpha}{48}$
  with probability at least $\nicefrac\beta6$.
  By \Cref{thm:replicable rounding},
  the rounded estimate $\wh p$ is $\nicefrac\rho3$-replicable and satisfies
  \[
    \abs{\wh p - p} \leq \nicefrac\alpha4
  \]
  with probability at least $\nicefrac\beta6$.
  From hereonforth,
  we condition on the success of this event.

  \paragraph{Correctness of \Cref{alg:replicable learner OWS}.}
  First suppose $\wh p < \nicefrac\alpha2$.
  Then
  \[
    p\leq \wh p + \nicefrac\alpha4 < \alpha.
  \]
  Then \Cref{alg:replicable learner OWS} always returns the all-zero hypothesis.
  In this case,
  the returned hypothesis only makes mistakes on at most $\alpha$ fraction of the population
  and we are content.

  Otherwise,
  suppose that $\wh p\geq \nicefrac\alpha2$.
  Thus
  \begin{align*}
    p &\geq \wh p - \frac\alpha4 \geq \frac\alpha4 \\
    \frac{n}{m} &\geq p - \frac{\rho\alpha}{16} \geq \frac\alpha8 \\
    n &\geq \frac\alpha8 m \\
    &= \frac{1152k^2}{\alpha^2\rho^2}\ln\frac6\beta.
  \end{align*}

  Let $F$ denote the CDF of the conditional distribution over the positively labeled samples.
  By \Cref{thm:replicable quantile est},
  the output of \rQuantileEst is an index $i^*\in [2^k]$
  such that with probability at least $\nicefrac\beta6$,
  \begin{align*}
    F(i^*) &\geq \frac\alpha2 - \frac\alpha4 \geq \frac\alpha4 \\
    F(i^*-1) &< \frac\alpha2 + \frac\alpha4 \leq \alpha.
  \end{align*}
  We proceed conditioning on the success of the call to \rQuantileEst.

  Since $F(i^*) \geq \nicefrac\alpha8$,
  the probability that $i_{j_1} > i^*$ is at most
  \[
    \left( 1 - \frac\alpha8\right)^n
    \leq e^{-\frac{n\alpha}8}
    \leq \nicefrac\beta6.
  \]
  The last inequality is due to $n\geq \nicefrac{\alpha m}8 \geq (\nicefrac4\alpha)\ln(\nicefrac6\beta)$.
  We proceed conditioning on the event $i_{j_1} \leq i^*$.

  So either $i_{j_1} = i^*$ and we know its string $\sigma^*$ and label $b^*$,
  or $i_{j_1} < i^*$ and \Cref{prop:compute forward} assures that
  we can obtain the string $\sigma^*$ and label $b^*$
  through forward computation
  \[
    (\sigma^*, b^*)
    \gets \ComputeForward(i^*, i_{j_1}, \sigma_{j_1}).
  \]

  Now,
  consider the hypothesis $h$ we output
  and its action on an input $(i, \sigma)$:
  If $i < i^*$,
  then $h$ always answers 0 so it is incorrect only on the positive labels.
  But this happens at most $F(i^*-1) < \alpha$ of the time.
  If $i \geq i^*$,
  $h$ is always correct.
  Hence the total population error is at most $\alpha$ as desired.

  We conditioned on three events,
  each of which has failure probability at most $\nicefrac\beta6$.
  Hence the total probability of failure is at most $\nicefrac\beta2$.

  \paragraph{Replicability of \Cref{alg:replicable learner OWS}.}
  By \Cref{thm:replicable rounding},
  the output $\wh p$ is $\nicefrac\rho3$ replicable.
  By \Cref{thm:replicable quantile est},
  the output $i^*$ is $\nicefrac\rho3$ replicable.
  From our analysis above,
  $i_{j_1} < i^*$ with probability at most $\nicefrac\beta6$ in each of two executions.
  Thus the total probability of outputting different classifiers is at most
  \[
    \frac\rho3 + \frac\rho3 + \frac{2\beta}6
    \leq \frac{2\rho}3 + \frac{\rho}9
    \leq \rho.
  \]
\end{pf}

\subsection{Proof of \texorpdfstring{\Cref{thm:main1}}{Theorem}}
We now restate and prove \Cref{thm:main1}.
\mainTheoremOne*

\begin{pf}[\Cref{thm:main1}]
    Combining the hardness of efficiently online learning \OWS (cf. \Cref{thm:online hardness OWS})
    with the efficient replicable learner of \Cref{thm:replicable learner OWS} completes the the proof of \Cref{thm:main1}.
\end{pf}

\section{Efficient Replicability and SQ Learning: Parities}
\label{sec:replicable but not sq}

\subsection{The Proof of \texorpdfstring{\Cref{lem:replicable affine parity}}{Lemma}}
\label{apx:replicable affine parity}
\begin{algorithm}[H]
\caption{Replicable Learner for Affine Parities $f(x) = w^\top x + b$ under the Uniform Distribution}\label{alg:replicable affine parity}
\begin{algorithmic}[1]
    \STATE $\mathtt{rAffParity}${(accuracy $\alpha$, replicability $\rho$, confidence $\beta$):}
    \STATE Draw a single sample $(x^{(0)}, f(x^{(0)}))$ where $x_0\sim \mcal U$.
    \STATE Draw $O(d \log(\nicefrac1{\rho\beta}))$ samples $(x, f(x))$ so that with probability at least $1-\beta$,
    we obtain $d$ linearly independent offsets $(z, y) := (x+x^{(0)}, f(x)+f(x^{(0)}))$,
    say $(z^{(1)}, y^{(1)}), \dots, (z^{(d)}, y^{(d)})$.
    \STATE Run Gaussian elimination to obtain the unique solution $w$ such that $w^\top z_i = y_i$ for each $i\in [d]$.
    \STATE Compute $b = f(x^{(0)}) + w^\top x^{(0)}$.
    \STATE Return $(w, b)$.
\end{algorithmic}
\end{algorithm}

We now repeat and prove \Cref{lem:replicable affine parity},
which states the correctness of \Cref{alg:replicable affine parity}.
\rAffineParity*

\begin{pf}[\Cref{lem:replicable affine parity}]
    For any $i\in [d]$,
    observe that
    \[
        y^{(i)}
        := f(x^{(i)}) + f(x^{(0)})
        := w^\top x^{(i)} + b + w^\top x^{(0)} + b
        = w^\top (x^{(i)} + x^{(0)})
        =: w^\top z^{(i)}
    \]
    Thus the dataset of offsets $\Set{(z^{(i)}, y^{(i)})}$ uniquely determines the linear function $w$.
    Having learned $w$,
    we can recover the value of $b$ by evaluating at any point in the original dataset,
    say at the first point.
\end{pf}

\subsection{Background on Parities and SQ}
\label{sec:background}

\paragraph{Statistical Queries} We start with some background on the SQ model.
\begin{defn}
[SQ Learning]
A concept class $\mathscr C$ with input space $\{0,1\}^d$ is learnable from statistical queries with respect to distribution $\mcal D$ if there is
a learning algorithm $\mcal A$ such that for any $c \in \mathscr C$ and any $\alpha > 0$, $\mcal A$ produces an $\alpha$-approximation
of $c$ from statistical queries; furthermore, the running time, the number of queries asked, and the
inverse of the smallest tolerance used must be polynomial in $d$ and $\nicefrac1\alpha$. 
\end{defn}

\begin{defn}
[SQ Hardness]
\label{def:sq-hard}
    Consider a concept class $\mathscr C$ with input space $\{0,1\}^d$.
    Fix a tolerance $\tau = \poly(\nicefrac1d)$ and accuracy $\alpha = O(1)$.
    We say that $\mathscr C$ is SQ-hard under distribution $\mcal D$ if any SQ algorithm requires $\omega(\poly(d))$ queries of tolerance at least $\tau$ to $\alpha$-learn $\mathscr C$.
\end{defn}

The standard way to show SQ hardness is by showing lower bounds on the so-called \emph{SQ
dimension} \citep{feldman2017statistical} of the corresponding problems 
(we omit the formal definition of SQ dimension as it is beyond the scope of the present work). 
Such lower bounds on this dimension establish lower bounds on the running time of any SQ algorithm
for the problem – not on its sample complexity.

\paragraph{PAC and SQ Learning Parities}
The standard class of parities $\mathsf{Parity}$ is given by the subset $\{f_{S,0} : S \subseteq [d]\}$ of $\mathsf{AffParity}$.
The standard algorithm for learning parity functions works by viewing
a set of $n$ labelled examples as a set of $n$ linear equations over the finite field with two elements $\mbb F_2$. Then, 
 Gaussian
elimination is used to solve the system and thus find a consistent parity function.
This algorithm is extremely brittle to noise: even for 
the uniform marginal distribution $\mathcal D = \mathcal U$, the problem of learning parities over $\{0,1\}^d$ does not belong to SQ.
In particular, a standard result is the following:

\begin{fact}
[\citep{kearns1998efficient}]
Even learning parities $\mathsf{Parity} \subset \mathsf{AffParity}$ with input space $\{0,1\}^d$ is SQ-hard under the uniform distribution $\mcal U$.
\end{fact}

\subsection{Gaussian Elimination is not Replicable }

In this section, we give a simple example where Gaussian elimination fails to be replicable.
\begin{prop}\label{prop:Gaussian elimination fails}
    For any $d\geq 3$
    and $S\sset [d]$ with $2\notin S\sset [d]$,
    the following holds:
    There is some $n_d\geq 1$,
    such that for any $n\geq n_d$,
    there exists a distribution $\mcal D$
    such that the $n$-sample Gaussian elimination algorithm for PAC-learning parities $f(x) = \sum_{i\in S} x_i$ in the realizable setting
    fails to be $\nicefrac1{17}$-replicable.
\end{prop}

\begin{pf}[\Cref{prop:Gaussian elimination fails}]
    Let $e_1, \dots, e_d$ be the standard basis of $\mbb Z_2^d$,
    $p\in (0, \nicefrac23]$,
    and consider the distribution $\mcal D$ such that
    \[
        \mcal D(e_1) = p, \qquad
        \mcal D(e_2) = \frac23-p , \qquad
        \mcal D(e_i) = \frac1{3(d-2)}, i \geq 3.
    \]
    The probability of not observing any $e_1$'s after $n$ samples is $(1-p)^n$.
    We set $p := 1-\sqrt[n]{\nicefrac12}\leq \nicefrac12$ so that $(1-p)^n = \nicefrac12$.
    
    For sufficiently large $n_d\in \N$,
    we observe all $e_i, i\geq 3$ with probability at least $\nicefrac34$ after $n$ samples.
    Thus with probability at least $\frac14$,
    we observe $e_1, e_3, \dots, e_d$,
    and with probability at least $\nicefrac14$,
    we observe $e_3, \dots, e_d$ but not $e_1$.
    In the first case,
    we fully recover $S$ since we are promised that $2\notin S$.    
    In the second case,
    our algorithm is unable to determine if $1\in S$ since we do not observe $e_1$
    and the best it can do is randomly guess.
    Thus the probability we output different classifiers is at least
    $2\cdot \nicefrac14\cdot \nicefrac14\cdot \nicefrac12 = \nicefrac1{16}$.
\end{pf}
\Cref{prop:Gaussian elimination fails} shows that Gaussian elimination fails in general to be replicable 
regardless of the number of samples requested.

\subsection{Replicably Learning Affine Parities Beyond the Uniform Distribution}
\label{apx:replicable affine parity non-uniform}
We restate our main result.
\begin{restatable}{thm}{rAffineParityNonUniform}\label{thm:replicable parities}
    For any dimension $d$, accuracy $\alpha$, confidence $\beta$ and replicability $\rho$,
    there exists some $n = \poly(d, \nicefrac1\alpha, \nicefrac1\rho, \log(\nicefrac1\beta))$
    such that
    there exists a monotone distribution $\mathcal{D}$ over $\{0,1\}^d$ which 
    \begin{enumerate}[(a)]
        \item Gaussian elimination is not $n$-sample replicable with probability $\Omega(1)$ under $\mcal D$ for the class $\mathsf{AffParity}$,
        \item there exists an $n$-sample $\rho$-replicable algorithm for $(\alpha,\beta)$-PAC learning the class $\mathsf{AffParity}$ with respect to $\mathcal{D}$ with $\mathrm{poly}(n)$ runtime, and
        \item even learning the class $\mathsf{Parity}$ is SQ-hard under the distribution $\mcal D$.
    \end{enumerate}
\end{restatable}

Before proving \Cref{thm:replicable parities},
we derive three useful lemmas. 
Fix a dataset size $n$ to be defined later.
We pick the distribution $\mcal D$ to be a Boolean product distribution $(p_1,...,p_d) \in [0,1]^d$ such that 
the first $d-1$ coordinates are unbiased $(p_i = 1/2)$ for $i \in [d-1]$ and the last coordinate to be highly biased towards 1. In particular, pick $p_d = \mcal D_d(\{1\}) = \sqrt[n]{\nicefrac12} \geq \nicefrac12$.
This distribution is monotone by construction.

\begin{lem}\label{lem:gaussian fails on product}
    The naive $n$-sample Gaussian elimination algorithm fails to be replicable with constant probability over the distribution $\mcal D = (p_1,...,p_d)$.
\end{lem}

\begin{pf}
    Consider two independent draws $S_1$ and $S_2$ from $\mcal D$ of size $n$.
    With constant probability $\nicefrac12\cdot \nicefrac12 = \nicefrac14$,
    $S_1$ will contain vectors that have only $1$'s in the last coordinate
    while $S_2$ will contain at least one vector with a zero in the last direction. 
    Then there are exactly two hypotheses that satisfy $S_1$,
    one that contains the last coordinate
    and one that does not.
    This can be seen by reducing to an instance of the $(d-1)$-dimensional parities problem
    obtained by ignoring the last coordinate
    and flipping the bits of the labels.
    
    On the other hand,
    for $n$ sufficiently large,
    the subset of vectors $S_2^{(1)}\sset S_2$ that have $1$'s in the last direction is satisfied by the same hypotheses as $S_1$.
    However,
    only one of the two hypotheses also satisfies the other subset $S_2^{(0)}\sset S_2$
    consisting of entries with a 0 in the last direction.

    Thus the candidate hypotheses that satisfy $S_1, S_2$ differ
    and the algorithm can at best output a random guess for $S_1$,
    which will be inconsistent with the output on $S_2$ with overall probability at least $\nicefrac12\cdot \nicefrac14 = \nicefrac18$.
\end{pf}

\begin{lem}
The decision tree complexity of $\mcal D$ is $\Theta(1)$.
\end{lem}

\begin{pf}
    Since $\mcal D$ is a product distribution, 
    the pmf only takes on $2$ different values 
    and thus has depth $\Theta(1)$. 
\end{pf}

\begin{lem}
\label{lem:closed}
The class of affine parities $\mathsf{AffParity}$ with input space $\{0,1\}^d$ is closed under restriction.
\end{lem}
\begin{pf}
Let $f$ be an arbitrary affine parity function, i.e., $f(x) = b + \sum_{i \in S} x_i$ for some $b \in \{0,1\}$ and $S \subseteq [d]$. We remark that the operator $+$ is in $\mathbb Z_2$. Consider an arbitrary direction $i \in [d]$ and arbitrary bit $b' \in \{0,1\}$. We will verify that
$f_{i = b'}$ remains in the class of affine parities.
If $i \notin S$, then $f_{i = b'} = f \in \mathsf{AffParity}$. 
Now, if $i \in S$ and $c = 0$, it holds that
$f_{i=b'}(x) = b + \sum_{i \in S\setminus \{i\}} x_i \in \mathsf{AffParity}$
and if $b' = 1$, it holds that
$f_{i=b'}(x) = (b+1) + \sum_{i \in S\setminus \{i\}} x_i \in \mathsf{AffParity}$.
Hence, affine parities are closed under restriction.
\end{pf}

We are now ready to prove \Cref{thm:replicable parities}.

\begin{pf}[\Cref{thm:replicable parities}]
    ~\paragraph{(a)}
    By \Cref{lem:gaussian fails on product}.
    
    \paragraph{(b)}
    We can employ the lifting algorithm of \Cref{thm:lift} to replicably and efficiently learn affine parities since (i) the decision tree complexity is constant, (ii) the distribution is monotone, (iii) affine parities are closed under restriction and (iv) we have an efficient replicable algorithm for affine parities with respect to the uniform distribution (cf. \Cref{lem:replicable affine parity}).
    
    \paragraph{(c)} If we consider the subclass $\mscr C' = \{f_{S,0} : S \subseteq [d-1]\}$, we have that $\E_{x \sim \mcal D}[f(x) g(x)] = 0$ for any pair of distinct functions $f,g \in \mscr C'$ due to the product structure and the unbiasedness of the first $d-1$ coordinates of $\mcal D$. 
    This implies the SQ-dimension of the class of parities is at least $2^{d-1} = \Omega(2^d)$ and so learning parities is SQ-hard under $\mcal D$. 
\end{pf}

\section{Lifting Replicable Uniform Learners}
\label{sec:lifting}

In this section, we will provide our general lifting framework for replicable learning algorithms that is needed in order to show \Cref{thm:replicable parities}.

\subsection{Preliminaries for Replicable Uniform Lifting}

We start this section with some definitions which are necessary to formally state our main result of \Cref{thm:lift}.

\begin{defn}
[Decision Tree (DT)]
\label{defn:decision tree}
A decision tree $T : \{0,1\}^d \to \mathbb R$ is a binary tree whose
internal nodes query a particular coordinate, and whose leaves are labelled by values. 
Each instance $x \in \{0,1\}^d$
follows a unique root-to-leaf path in $T$: 
at any internal node, it follows either the left or
right branch depending on the value of the queried coordinate, until a leaf is reached and its value $T(x)$
is returned.
\end{defn}

\begin{defn}
[Decision tree distribution]
We say that a distribution $\mcal D$ over $\{0,1\}^d$ 
is representable by a depth-$\ell$ decision tree, if its pmf is computable by a depth-$\ell$ decision tree $T$. Specifically, each
leaf $t$ is labelled by a value $p_t$, so that $\mcal D(x) = p_t$ for all $x \in t$.
This means that the conditional
distribution of all points that reach a leaf is uniform.
\end{defn}

Let us consider a distribution $\mcal D$ over $\{0,1\}^d$. 
The definitions above suggest a natural measure of the complexity of $\mcal D$:
The \emph{decision tree complexity}.
\begin{defn}
[Decision Tree Complexity]
\label{def:DT} 
The {decision tree complexity} of a distribution $\mathcal{D}$ over $\{0,1\}^d$ is the smallest integer $\ell$ such that its probability mass function (pmf) can be represented by a depth-$\ell$ decision tree. 
\end{defn} 

Next, we define a useful structured family of distributions.
Let the binary relation $\succeq$ denote pointwise comparison.
\begin{defn}
[Monotone Distribution]
\label{def:monotone}
A probability distribution $\mathcal{D}$ over $\{0,1\}^d$ is called monotone if for any $x \succeq y$ it holds that $\mathcal D(x) \geq \mathcal{D}(y).$   
\end{defn}

Just as in \citet{blanc2023lifting},
for arbitrary non-monotone probability measures, 
our algorithm requires access to a \emph{conditional sampling oracle} (cf. \Cref{def:conditional sampler}) defined below.
\begin{defn}[Conditional Sampling Oracle; \citealp{blanc2023lifting}]\label{def:conditional sampler}
    A conditional sampling oracle for a distribution $\mcal D$ over $\Set{0, 1}^d$ proceeds as follows: 
    suppose we condition on some subset $I$ of the $d$ variables having a fixed value $b\in \Set{0, 1}^I$. In that case, the oracle generates a sample from 
    the true conditional distribution, i.e., draws a sample $x_{-I} \sim \mcal D(\cdot \mid x_I = b)$.
\end{defn}

\begin{defn}
\label{def:expectation-leaves}
    For a tree $T$ and leaves $\ell \in T$, $\E_{\ell \in T} f(\ell) := \sum_{\ell \in T} 2^{-|\ell|} f(\ell)$, where $|\ell|$ is the depth of the path to reach leaf $\ell.$
\end{defn}
As an example of this notation, note that
\[
    \E_{\ell \in T} \left[ \E_{x \sim \mcal U^d} [f(x) | x \in \ell] \right] 
    = \E_{\ell \in T} \left[ \sum_{x \in \ell} \frac{f(x) \mcal U(x)}{ \mcal U(\ell) } \right],
\]
where $\mcal U(\ell)$ is the uniform mass of the leaf $\ell$, i.e., $\mcal U(\ell) = 2^{-n} \cdot 2^{n-|\ell|}$. This implies that $\E_{\ell \in T}[\E_{x \sim \mcal U^d} [f(x) | x \in \ell]] = \E_{x \in \mcal U^d} f(x),$ since the set of leaves partitions the set $\{0,1\}^d$. 

\subsection{Main Result: Lifting Replicable Uniform Learners}
We now restate our main result \Cref{thm:lift}.
For the sake of presentation,
we defer its proof to \Cref{apx:lifting}.
\rLiftUniformLearners*

We emphasize that our result for monotone distributions $\mcal D$ only requires sample access to $\mcal D$. For arbitrary non-monotone probability measures, our algorithm requires
access to a conditional sampling oracle (cf. \Cref{def:conditional sampler})

\section{Proof of \texorpdfstring{\Cref{thm:lift}}{Theorem}}\label{apx:lifting}

\subsection{Preliminaries}

We start this section with some useful definitions, coming from the work of \cite{blanc2023lifting}.

\begin{defn}
[Weighting function of distribution] 
\label{def:weight}
Let $\mcal D$ be an arbitrary distribution over $\{0,1\}^d$.
We define the weighting function
$f_{\mcal D}(x) = 2^d \mcal D(x).$
\end{defn}
Remark that the scaled pmf satisfies
\[
    \E_{x\sim \mcal U} [f_{\mcal D}(x)]
    = \sum_{x\in \Set{0, 1}^d} 2^{-d}\cdot 2^{d} \mcal D(x)
    = 1.
\]

For $x\in \Set{0, 1}^d$ and $i\in [d]$,
we write $x^{\sim i}$ to denote the binary vector obtained from $x$
by flipping its $i$-th bit.
\begin{defn}
[Influence of Variables on Distributions]
\label{def:inf} 
    Let $\mathcal{D}$ be a distribution over $\{0,1\}^d$ and  $f_\mathcal{D}(x) = 2^d \cdot \mathcal{D}(x)$ be its pmf scaled up by the domain size.  The {influence} of a coordinate $i\in [d]$ on a distribution $\mathcal{D}$ over $\{0,1\}^d$ is the quantity 
    \[ 
        \Infl_i(f_\mathcal{D}) 
        := \E_{x\sim \mcal U} \left[ \abs{f_{\mcal D}(x) - f_{\mcal D}(x^{\sim i})} \right]
    \]
    We also define the \emph{total influence} $\Infl(f_\mathcal{D}) = \sum_{i \in [d]} \Infl_i(f_\mathcal{D})$.
\end{defn}
{Intuitively,
the influence of a variable measures how far that variable is from the uniform distribution.}

Suppose $f_{\mcal D}$ is computable by a depth-$\ell$ decision tree $T$.
Then we can write for any $x\in \Set{0, 1}^d$
\[
    \mcal D(x)
    = \sum_{\ell\in T} p_\ell \cdot \ones\Set{x\in \ell}.
\]
where the sum is taken over the leaves of $T$
and $p_\ell:= \mcal D(y)$ is the (same) mass $\mcal D$ assigns to every $y\in \ell$.
Then we have
\begin{align*}
    \Infl_i(f_{\mcal D})
    &:= \E_{x\sim \mcal U} \abs{f_{\mcal D(x)} - f_{\mcal D}(x^{\sim i})} \\
    &= \sum_{x\in \Set{0, 1}^d} \abs{\mcal D(x) - \mcal D(x^{\sim i})} \\
    &= \sum_x \abs*{\sum_{\ell\in T} p_\ell \left( \ones\Set{x\in \ell} - \ones\Set{x^{\sim i}\in \ell} \right)}.
\end{align*}
If $i$ is not queried by any internal node of $T$,
then $x, x^{\sim i}$ always belongs to the same leaf
and this value is 0.
Otherwise,
we can trivially bound this value by 2 using the triangle inequality.
This discussion leads to the following observation.
\begin{fact}\label{fact:total infl bound}
    {If $f_{\mcal D}$ is computable by a depth-$\ell$ decision tree,
    then its total influence is at most
    \[
        \Infl(f_\mathcal{D})
        \leq 2\ell.
    \]}
\end{fact}

For monotone distributions over the Boolean hypercube, influences have a convenient form:
\begin{prop}[Lemma 6.2 in \citet{blanc2023lifting}]\label{prop:monotone influence}
    If the distribution $\mcal D$ over $\{0,1\}^d$ is monotone, 
    then for any $i \in [d]$
    \[
        \Infl_i(f_{\mathcal D}) = \E_{\mathcal D}[x_i]\,.
    \]
\end{prop}

\begin{defn}
[Restrictions]
\label{def:restriction}
Given a
sequence of (coordinate, value) pairs $\pi = \{(i_1, b_1),...,(i_k, b_k)\}$, we use $x_\pi$ to represent $x$ with the
coordinates in $\pi$ overwritten/inserted with their respective values.
For a function $f : \{0,1\}^d \to \mathbb R,$ we let $f_\pi$ be the function that maps $x$ to $f(x_\pi)$.
\end{defn}

\begin{defn}
[Everywhere $\tau$-influential; Definition 4 in \citet{blanc2022properly}] 
\label{def:influential}
    For any function $f : \{0,1\}^d \to \mathbb  R$, influence threshold $\tau > 0$ and decision tree $T : \{0,1\}^d \to \mathbb R$, we say that $T$ is everywhere $\tau$-influential with respect to some $f$ if 
    for every internal node $v$ of $T$, 
    we have 
    \[
    \Infl_{i(v)}(f_v) \geq \tau\,.
    \]
    Here $i(v)$ denotes the variable queried at $v$
    and $f_v$ denotes the restriction of $f$ by the root-to-$v$ path in $T$,
    i.e. the variables in the path are fixed to the corresponding internal vertex values.
\end{defn}

\subsection{Replicable Proper Learner for Decision Tree distributions}
Our algorithm that achieves the bounds 
of~\Cref{thm:lift} proceeds in a two-stage manner: we first replicably learn the decision tree structure of $\mathcal{D}$ and then use the replicable uniform-distribution learner to learn $f$ restricted to each of the leaves of the tree. To carry out the first stage, we give an algorithm that replicably learns the {optimal} decision tree decomposition of a distribution $\mathcal{D}$: 

\begin{thm}
[Replicably Learning DT distributions]
\label{thm:learn-DT} 
    {Fix $\alpha, \rho\in (0, 1)$ and $\beta\in (0, \nicefrac\rho3)$.}
    Let $\mathcal D$ be a distribution over $\Set{0, 1}^d$ that is representable by a depth-$\ell$ decision tree.
    There is an algorithm that
    \begin{enumerate}[(a)]
        \item is $\rho$-replicable with respect to $\mathcal D$,
        \item returns a depth-$\ell$ tree representing a distribution $\mathcal D'$
        such that $\TV(\mathcal D,\mathcal D') \le \alpha $
        with probability at least $1-\beta$ over the draw of samples,
        \item and has running time and sample complexity
        \[
            {
            N
            = \poly(d, \nicefrac1\alpha, \nicefrac1\rho, \log(\nicefrac1\beta))\cdot (\nicefrac{2\ell}{\alpha})^{O(\ell)}
            }.
        \]
    \end{enumerate}
    For monotone distributions, the algorithm only uses random samples from $\mathcal D$, and for general distributions, it uses subcube conditional samples (cf. \Cref{def:oracle}).
\end{thm}

\begin{pf}
[\Cref{thm:learn-DT}]
The proof of this result follows directly by applying either \Cref{lemma:influence monotone} (for monotone distributions) or \Cref{lemma:influence general} (for general distributions with subcube conditional sample access) to \Cref{thm:influence implies tree learning}, which are provided right after.
\end{pf}

Essentially, making the decision tree learning routine replicable boils down to estimating influences on $\mathcal D$ in a replicable manner. This is the content of the upcoming results.

\begin{thm}
[Replicable Influence $\Rightarrow$ Replicable Decision Tree]
\label{thm:replicable influence to replicable decision tree}
\label{thm:influence implies tree learning}
Let $\mathcal D$ be a distribution that is representable by a depth-$\ell$ decision tree.
For any $\rho' \in (0,1)$ assume access to a $\rho'$-replicable algorithm $\rInflEst$ for estimating influences of distributions over $\{0,1\}^d$ using $\poly(d, \nicefrac1\alpha, \nicefrac1{\rho'}, \log(\nicefrac1\beta))$ samples and runtime.
Then, for any $\rho \in (0,1)$, there is an algorithm $\rBuildDT$ (cf. \Cref{algorithms:universal1}) that 
\begin{enumerate}[(a)]
    \item is $\rho$-replicable with respect to $\mathcal D$,
    \item returns a depth-$\ell$ tree representing a distribution $\mathcal D'$
such that $\TV(\mathcal D,\mathcal D') 
    \le \alpha $, with probability at least $1-\beta$ over the draw of samples,
    \item and has running time and sample complexity
    {
    \begin{align*}
        M 
        &= \poly\left( d, \frac1{\alpha}, \frac{d(\nicefrac{2\ell}\alpha)^{O(\ell)}}{\rho},
        \log \frac{d(\nicefrac{2\ell}\alpha)^{O(\ell)}}\beta \right)\cdot d\cdot (\nicefrac{2\ell}\alpha)^{O(\ell)} \\ 
        &\qquad + O\left( \frac{(\nicefrac{2\ell}\alpha)^{O(\ell)}}{\alpha^2 \rho^2} \log\frac{(\nicefrac{2\ell}\alpha)^{O(\ell)}}\beta \right)\cdot (\nicefrac{2\ell}\alpha)^{O(\ell)} \\
        &= \poly(d, \nicefrac1\alpha, \nicefrac1\rho, \log(\nicefrac1\beta))\cdot (\nicefrac{2\ell}{\alpha})^{O(\ell)}.
    \end{align*}
    }
\end{enumerate}
\end{thm}

\begin{algorithm}[ht!]
  \caption{Replicably Building Decision Tree for $\mathcal D$}
    \begin{algorithmic}[1]
    \STATE \rBuildDT:
    \STATE \textbf{Input:} Access to $\mathcal{D}$ and $f_{\mcal D}$ as in \Cref{def:weight}, 
    restriction $\pi$ (cf. \Cref{def:restriction}), 
    access to $\rInflEst$, 
    depth $\ell$, 
    influence threshold $\tau$, 
    accuracy $\alpha$,
    confidence $\beta$,
    replicability $\rho$.
    
    \STATE \textbf{Output:}
    A decision tree $T$ that minimizes 
    $\E_{t \in T}[\Infl((f_{\mathcal D})_t)]$, where $t \in T$ is the collection of leaves (each corresponding to some restriction) of $T$
    among all depth-$\ell$, everywhere $\tau$-influential (cf. \Cref{def:influential}) trees.
    \FOR {$i\in [d]$}
        \STATE
        $
            \rInflEst[(f_\mathcal{D})_\pi, i]
            \gets \rInflEst\left(
            (f_\mathcal{D})_\pi, 
            i, 
            \min(\nicefrac\tau4, 
            \nicefrac\alpha{2d}), 
            \frac\beta{2d (\nicefrac{2\ell}\alpha)^{O(\ell)}}, 
            \frac\rho{2d (\nicefrac{2\ell}\alpha)^{O(\ell)}} \right)
        $
    \ENDFOR
    \STATE Let $S \subseteq [d]$ 
    be the set of variables $i$ so that
    $
        {
        \rInflEst[(f_\mathcal{D})_\pi, i]\geq \nicefrac{3\tau}4.
        }
    $
    \IF {$S = \emptyset$ or $\ell = 0$}
        \STATE \COMMENT {replicably estimate $p_\pi := 2^{\card\pi} \Pr_{x \sim \mathcal D}[x \mathrm{~is~consistent~with~} \pi]$ \qquad (cf. \Cref{thm:replicable rounding})}
        \STATE $\hat p_\pi \gets \alpha/2$-accurate, 
        $\rho/[2 (\nicefrac{2\ell}\alpha)^{O(\ell)}]$-replicable, 
        $\beta/[2 (\nicefrac{2\ell}\alpha)^{O(\ell)}]$-confident
        estimate of $p_\pi$
        \RETURN a new leaf node with value {$\hat p_\pi$} \label{line:leaf return value}
    \ELSE
        \FOR {$i \in S$}
            \STATE construct the tree $T_i$ with
            \STATE $\mathrm{root}(T_i) \gets x_i$
            \STATE $\mathrm{leftSubtree}(T_i) \gets \rBuildDT(\mathcal D, \pi \cup \{x_i = -1\}, \ell-1, \tau, \alpha, \beta, \rho)$
            \STATE $\mathrm{rightSubtree}(T_i) \gets \rBuildDT(\mathcal D, \pi \cup \{x_i = 1\}, \ell-1, \tau, \alpha, \beta, \rho)$
        \ENDFOR
        \STATE \COMMENT Among $\{T_i\}_{i \in S}$, return the one that minimizes the estimated total influence
        \STATE \COMMENT 
        $
            \E_{t \in T_i}\left[ \sum_{j\in [d]} \rInflEst[(f_\mathcal{D})_t, j] \right]
        $
        where $(f_{\mcal D})_t = (f_{\mcal D})_{\pi(t)}$ 
        \STATE \COMMENT for the restriction $\pi(t)$ corresponding to path from the root the leaf $t$
        \STATE \COMMENT Let $g(t) := \sum_{j\in [d]} \rInflEst[(f_\mathcal{D})_t, j]$
        \STATE {$i^* \gets \argmin_{i\in S}\E_{t\in T_i} [g(t)] 
        = \sum_{t\in T_i} 2^{-\card{t}} g(t)
        = \sum_{x\in \Set{0, 1}^d} 2^{-d} g(t)\ones\Set{x\in t}$}
        \RETURN $T_{i^*}$
    \ENDIF
    \end{algorithmic}
\label{algorithms:universal1}
\end{algorithm}

Before proving \Cref{thm:replicable influence to replicable decision tree},
we state Theorem 3 of \citet{blanc2023lifting},
from which the correctness of \Cref{algorithms:universal1} follows.
Let $\mathtt{BuiltDT}$ be the algorithm obtained from \rBuildDT (cf. \Cref{algorithms:universal1}) with the following adjustments.
\begin{enumerate}[(i)]
    \item Replace the replicable influence estimator \rInflEst
with any (possibly non-replicable) influence estimator and
    \item replace the replicable mean estimator of $p_\pi$ with a simple (non-replicable) sample mean.
\end{enumerate}

\begin{prop}
[Theorem 3 and Claim 5.5 in \citet{blanc2023lifting}]
\label{prop:influence to decision tree}
Fix $\alpha, \beta\in (0, 1)$.
Let $\mcal D$ be a distribution that is representable by
a depth-$\ell$ decision tree. 
Given oracle access to an estimator for influences,
the algorithm $\mathtt{BuildDT}$ 
{with the choice of $\tau = \nicefrac\alpha{8\ell^2}$}
returns a depth-$\ell$ tree representing a distribution
$\mcal D'$
such that $\mathrm{TV}(\mcal D, \mcal D') \leq \alpha$ {with probability at least $1-\beta$}. 
If the oracle terminates in unit time,
the running time of the algorithm is $d\cdot (\ell/\alpha)^{O(\ell)}$.
\end{prop}

\begin{rem}
    A few remarks are in order regarding \Cref{algorithms:universal1}.
\label{rem:tree alg params}
    \phantom{placeholder}
    \begin{enumerate}[1.]
        \item \citet{blanc2023lifting} identify the task of learning of a decision tree for a distribution $\mcal D$
        with learning its \emph{scaled} pmf $f = 2^d \mcal D$.
        In order to remain consistent with their convention,
        the values at the tree leaves produced by our replicable decision tree algorithm (cf. \Cref{algorithms:universal1}) also correspond to the scaled pmf.
        Note that we do not directly use the values at the leaves and hence this convention is immaterial to our application.
        \item Estimating the influences to an accuracy of $\nicefrac\tau4$ ensures that any variables in $S$ has influence at least $\nicefrac\tau2$
        and every variable with influence at least $\tau$ is captured in $S$.
        Estimating the influences to an accuracy of $\nicefrac\alpha{2d}$ ensures that the estimated total influences are accurate up to error at most $\alpha$.
        Note that the computation for $\E_{t\in T_i} [g(t)]$ (cf. \Cref{def:expectation-leaves}) does not incur additional error since the weights of the sum are appropriately chosen.
        
        \item $p_\pi$ is the mean of a $[0, 2^{\card \pi}]$-bounded random variable.
        By an Hoeffding bound and \Cref{thm:replicable rounding},
        consuming $\poly(2^\ell, \nicefrac1\alpha, \nicefrac1{\rho'}, \log(\nicefrac1{\beta'}))$ samples from $\mcal D$ suffices to estimate this quantity
        up to $\alpha/[2\cdot 2^{\card\pi}]$-accuracy,
        $\rho'$-replicability, 
        and $\beta'$-confidence.
        
        \item {Taking into account the estimation error of influences,
        each variable in $S$ has influence at least $\nicefrac\tau2$.
        For any depth-$\ell$ decision tree,
        the sum of all variable influences is at most $2\ell$ (cf. \Cref{fact:total infl bound}).
        For any restriction $\pi$,
        $(f_{\mcal D})_\pi$ is a depth-$\ell$ decision tree
        and thus has at most $\nicefrac{4\ell}{\tau}$ variables of influence at least $\nicefrac\tau2$.
        For the choice of $\tau = \nicefrac\alpha{8\ell^2}$,
        there can be at most $(\nicefrac{4\ell}{\tau})^\ell = (\nicefrac{2\ell}\alpha)^{O(\ell)}$ recursive calls within \Cref{algorithms:universal1}.}
        
        Each recursive call estimates $d$ influences
        and at most $1$ mean of a bounded random variable.
        All in all,
        we need to replicably estimate $d\cdot (\nicefrac{2\ell}\alpha)^{O(\ell)}$ variable influences
        and at most $(\nicefrac{2\ell}\alpha)^{O(\ell)}$ bounded means.
    \end{enumerate}

\end{rem}

We are now ready to prove \Cref{thm:replicable influence to replicable decision tree}.
\begin{pf}[\Cref{thm:replicable influence to replicable decision tree}]
    By assumption,
    for any $\rho'\in (0, 1)$,
    \rInflEst is $\rho'$-replicable and returns $\alpha$-accurate estimates for influences with confidence $\beta$
    using $\poly(d, \nicefrac1\alpha, \log(\nicefrac1\beta), \nicefrac1{\rho'})$ samples and runtime.
    
    The replicability of \Cref{algorithms:universal1} follows directly from the replicability of the influence estimators
    and the replicability of the standard rounding scheme (cf. \Cref{thm:replicable rounding}).
    By \Cref{rem:tree alg params}, 
    it suffices to call the replicable influence estimator
    and the replicable mean estimator for $p_\pi$
    with replicability parameter $\rho' = \rho / [2d (\nicefrac{2\ell}\alpha)^{O(\ell)}]$ for the union bound
    and similarly for the confidence parameter. 
    The sample complexity follows from our remark above.
    
    Let us now condition on the event that each call of $\rInflEst$ and estimation of expected total influence is successful. 
    Conditioned on this event, the correctness of the algorithm $\rBuildDT$ directly follows from \Cref{prop:influence to decision tree}.
    In particular, if we condition on the event that each call to $\rInflEst$
    and each replicable estimate of $p_\pi$ is successful, our algorithm reduces to the $\mathtt{BuildDT}$ algorithm of \citet{blanc2023lifting}.

\end{pf}

\subsection{Replicable Influence Estimator for Monotone Marginals}
In this section, we provide the main subroutine of our replicable DT distribution learning algorithm.
We begin with a replicable influence estimator for monotone distributions.

\begin{lem}
[Replicable Influence Estimator for Monotone Marginals]
\label{lemma:influence monotone}
Fix $i \in [d]$.
For any $\alpha,\beta, \rho \in (0,1)^3$, there is an
efficient $\rho$-replicable algorithm $\rInflEst(\mcal D, i, \alpha, \beta, \rho)$ such that given an unknown monotone distribution $\mathcal D$ over $\{0,1\}^d$, 
 computes an estimate of 
$\Infl_i(f_{\mathcal D})$
up to accuracy $\alpha$ with probability at least $1-\beta$ using 
$O(\alpha^{-2} \rho^{-2} \log(\nicefrac1\beta))$ time and random samples from $\mathcal D$.
\end{lem}

\begin{pf}[\Cref{lemma:influence monotone}]
    By \Cref{prop:monotone influence},
    since $\mathcal D$ is monotone, 
    we know that $\Infl_i(f_{\mathcal D}) = \E_{\mathcal D}[x_i]$. 
    Thus we can replicably estimate this expectation with $O(\alpha^{-2} \rho^{-2} \log(\nicefrac1\beta))$ samples and sample-polynomial runtime through replicable query rounding (cf. \Cref{thm:replicable rounding}).
\end{pf}

\subsection{Replicable Influence Estimator for Arbitrary Distributions}

We further design a replicable influence estimator for arbitrary distributions $\mcal D$ over $\{0,1\}^d$ using subcube
conditional sampling. 

\begin{defn}
\label{def:oracle}
A (subcube) conditional sampling oracle for distribution $\mcal D$ over $\{0,1\}^d$
receives as input a subset (subcube) $S \subseteq \{0,1\}^d$ 
and generates a sample from the conditional distribution
\[
    \mcal D_S(x)
    := \mcal D(x \mid S)
    = \frac{\mcal D(x) \ones\{x \in S\}}{\mcal D(S)}.
\]
\end{defn}

This conditional distribution is essentially the truncated distribution $\mcal D_S$ with truncation set $S$. The subcube conditioning oracle is well studied by prior work in distribution testing and learning \citep{canonne2015testing,canonne2021random,fotakis2022perfect,fotakis2020efficient,gouleakis2017faster}.

\begin{algorithm}[ht!]
  \caption{Replicable Influence Estimation for Arbitrary Distributions via Conditional Sampling}
    \begin{algorithmic}[1]
    \STATE \rInflEst:
    \STATE \textbf{Input:} 
    Distribution $\mcal D$ over $\{0,1\}^d$ and $f_{\mcal D}$ as in \Cref{def:weight},
    coordinate $i \in [d]$,
    access to conditional sampling oracle for $\mcal D$,
    desired accuracy $\alpha$,
    desired confidence $\beta$,
    desired replicability $\rho$.
    
    \STATE \textbf{Output:}
    A estimate $v$ of $\Infl_i(f_{\mcal D})$ such that $|v - \Infl_i(f_{\mcal D})| \leq \alpha$ with probability $1-\beta$.

    \STATE $n \gets O(\alpha^{-2} \rho^{-2}\log(\nicefrac1\beta))$
    \FOR {$j \gets 1,...,n$}
        \STATE Draw a fresh random sample $x^{(j)} \sim \mcal D$
    
        \STATE $S^{(j)} \gets \{x^{(j)} \} \cup \{(x^{(j)})^{\sim i}\}$, where $(x^{(j)})^{\sim i}$ is $x^{(j)}$ with its $i$-th coordinate flipped
    
        \STATE Draw $O(\alpha^{-2})$ independent samples from the conditional on $S^{(j)}$ distribution $\mcal D(\cdot | S^{(j)})$ 
        \STATE Let $p^{(j)} \in [0,1]$ be the fraction that we observe $x^{(j)}$
    
        \STATE $q^{(j)} \gets |p^{(j)} - (1-p^{(j)})|$
    \ENDFOR
    \STATE $q \gets \frac{1}{n} \sum_{j \in [n]} q^{(j)}$
    
    \RETURN $v \gets \mathtt{rRound}(q, O(\alpha\rho), \rho)$ \qquad (see \Cref{thm:replicable rounding} with $T =1$)    
    \end{algorithmic}
\label{algorithms:subscube}
\end{algorithm}

An adaptation of the proof of Proposition 6.5 in \citet{blanc2023lifting} combined with the replicability properties of \Cref{thm:replicable rounding} gives the following result, which we provide for completeness but do not use in our applications. The algorithm is presented \Cref{algorithms:subscube}.
\begin{lem}
[Replicable Influence Estimator for Arbitrary Marginals]
\label{lemma:influence general}
    Let $\mcal D$ be a distribution over $\{0,1\}^d$, $f_{\mcal D}$ as in \Cref{def:weight}, and, assume sample access to a subcube conditional oracle as in \Cref{def:oracle}. 
    For any $\alpha, \beta, \rho\in (0, 1)$ and $i \in [d]$,
    \Cref{algorithms:subscube} is $\rho$-replicable and, given $O(\alpha^{-4} \rho^{-2} \log(\nicefrac1\beta))$ subcube conditional samples from $\mcal D$, it computes in sample-polynomial time an estimate $v$ that satisfies
    $|v - \Infl_i(f_{\mcal D})| \leq \alpha$
    with probability at least $1-\beta$.
\end{lem}

As a proof sketch,
let us call the above algorithm $\mathtt{ReplInfEst}(\mcal D, i, \alpha)$.
The correctness of the algorithm follows from the observation that for any distribution $\mcal D$, coordinate $i$ and $\alpha \in (0,1)$, for any execution $j \in [n]$, it holds that $|\E q^{(j)} - \mathrm{Infl}_i(f_{\mcal D}) | \leq \alpha$. Hence to obtain a high probability estimator,
it suffices to obtain $O(\alpha^{-2}\log(\nicefrac1\beta))$ copies of $q^{(j)}$ and take the average $q$. Finally, to make the algorithm replicable, it suffices to estimate the average $q$ replicably. This can be accomplished at an extra cost of order $\nicefrac1{\rho^2}$ using the standard rounding routines (cf. \Cref{thm:replicable rounding}). Since each iteration requires $\nicefrac1{\alpha^2}$ subcube samples, the sample complexity of our algorithm follows.

\subsection{Useful Subroutines \& Results}
In this section, we provide a set of useful results that we will use in our proofs.

\begin{prop}[Corollary D.21 in \citet{esfandiari2023replicable}]\label{cor:replicable mass estimation}
  Let $\alpha, \rho\in (0, 1)$ and $\beta\in (0, \nicefrac\rho3)$.
  There is a $\rho$-replicable algorithm $\rFiniteDistrEst(\mcal D, \alpha, \beta, \rho)$
  that outputs parameter estimates $\bar p$ for a finite distribution $\mcal D$ of support size $N$
  such that
  \begin{enumerate}[(a)]
        \item $\abs{\bar p^{(i)} - p^{(i)}}\leq \alpha$
        for every $i\in [N]$
        with probability at least $1-\beta$.
        \item $\bar p^{(i)}\geq 0$ for all $i\in [N]$.
        \item $\sum_i \bar p^{(i)} = 1$.
  \end{enumerate}
  Moreover,
  the algorithm has sample complexity
  \[
    m
    = O\left( \frac{\ln\nicefrac1\beta + N}{\alpha^2 (\rho-\beta)^2} \right)
    = O\left( \frac{N}{\alpha^2 \rho^2}\log\frac1\beta \right)
  \]
  and $\poly(m)$ time complexity.
\end{prop}

\begin{prop}[\citep{janson2018tail}]\label{prop:multiple coupon collector}
  Let $Y_i$ be a geometric variable with success rate $q$.
  Then $Y := \sum_{i=1}^m Y_i$ is the number of draws until we obtain $m$ successes.
  Then
  \begin{align*}
    \P\Set*{Y\geq \lambda \frac{m}{q}} &\leq \exp(1-\lambda).
  \end{align*}
\end{prop}
In other words,
it suffices to perform $O(\nicefrac{m}{q}\log(\nicefrac1\beta))$ Poisson trials
before succeeding $m$ times with probability at least $1-\beta$.

\begin{prop}[Replicable Boosting of Success Probability]\label{prop:replicable boosting}
    Let $\alpha, \rho\in (0, 1)$,
    $\beta\in (0, \nicefrac\rho3)$,
    and $\Delta \geq 0$.
    Suppose $\mcal A$ is a $\rho'$-replicable $(\alpha + \Delta, \nicefrac1{2})$-PAC learner for the concept class $\mscr C$
    under a fixed distribution $\mcal D$ over $\Set{0, 1}^d$
    that uses $m(\alpha, \rho') = \poly(d, \nicefrac1\alpha, \nicefrac1{\rho'})$ samples
    for any $\rho'\in (0, 1)$.
    There is a $\rho$-replicable $(2\alpha+\Delta, \beta)$-PAC learner $\rBoost_{\mcal A}(\mcal D, \alpha, \beta, \rho)$ for $\mscr C$ under $\mcal D$
    for any $\alpha, \beta, \rho\in (0, 1)$
    that has sample complexity
    \[
        m\left( d, \frac\rho{2\log(\nicefrac1\beta)} \right)\cdot O\left( \log\frac{1}{\beta} \right)
        + O\left( \frac{\log^2 (\nicefrac1\beta)}{\alpha^2\rho^2} \log\frac{\log (\nicefrac1\beta)}\beta \right)
        = \poly\left( d, \nicefrac1\alpha, \nicefrac1\rho, \log(\nicefrac1\beta) \right).
    \]
    and sample-polynomial running time.
\end{prop}

\begin{pf}
    Let $\rBoost_{\mcal A}$ be the algorithm that runs $\mcal A$ for $n = O(\log\nicefrac1\beta)$ times (each with $m = m(\nicefrac1\alpha, \nicefrac\rho{2n})$ samples),
    replicably estimates the population error of each hypothesis,
    and outputs the hypothesis which has the lowest estimated error.

    Running $\mcal A$ for $n = O(\log\nicefrac1\beta)$ times with $m$ samples per run
    guarantees that each execution is $\nicefrac{\rho}{2n}$-replicable
    so the first step is $\nicefrac\rho2$-replicable.
    Moreover,
    at least one of the hypotheses we output is $(\alpha+\Delta)$-close to the optimal hypothesis
    with probability at least $1-\nicefrac\beta2$.

    By an Hoeffding bound on the empirical error of a hypothesis,
    we can estimate the population error of a hypothesis to an accuracy of $\nicefrac{\alpha \rho}{4\cdot 2n}$ and confidence $\nicefrac\beta{2n}$
    using $O(n^2 \alpha^{-2} \rho^{-2} \log (\nicefrac{n}\beta))$ samples.
    By \Cref{thm:replicable rounding},
    we can make this estimate $\nicefrac\rho{2n}$-replicable at the cost of reducing the accuracy to $\alpha$
    while maintaining the confidence parameter $\nicefrac\beta{2n}$.

    It follows that the output hypothesis is $\rho$-replicable
    and with probability at least $1-\beta$,
    its population error is at most $2\alpha+\Delta$.
\end{pf}

\subsection{Proof of \texorpdfstring{\Cref{thm:lift}}{Theorem}}

In this section, we show how to lift replicable algorithms that learn over the uniform distribution to replicable algorithms
that learn with respect to arbitrary distributions, where the sample complexity and running time depend on the decision tree complexity of the target distribution.
Let $\mcal D(\cdot\mid T)$ denote the conditional distribution on leaves,
i.e.,
$\mcal D(t\mid T) := \sum_{x\in t} \mcal D(x)$.

\begin{algorithm}[ht!]
  \caption{Replicable Lifting}
    \begin{algorithmic}[1]
    \STATE \rLift
    \STATE \textbf{Input:} 
    decision tree $T$ (from $\rBuildDT$),
    algorithm $\mathcal{A}$
    for $\rho'$-replicable $(\alpha',\beta')$-learning under the uniform distribution using $m(\alpha', \rho', \beta')$ samples,
    {distribution $\mcal D$},
    accuracy $\alpha$,
    confidence $\beta$,
    replicability $\rho$.
    \STATE \textbf{Output:}
    A hypothesis $h : \{0,1\}^d \to \{0,1\}$
    \STATE
    \STATE $M_1 \gets \poly(2^\ell, \nicefrac1\alpha, \nicefrac1\rho, \log(\nicefrac1\beta))$
    \STATE $M_2 \gets m(\nicefrac\alpha6, \nicefrac\rho{2^\ell}, \nicefrac16)\cdot 
    \poly(d, 2^\ell,\nicefrac1\alpha, \nicefrac1\rho, \log(\nicefrac1\beta))$
    \STATE \COMMENT Estimate leaf distribution (cf. \Cref{cor:replicable mass estimation})
    \STATE $\hat{\mcal D}(\cdot\mid T) \gets \rFiniteDistrEst(\mcal D(\cdot\mid T), \nicefrac\alpha{12\cdot 2^\ell}, \nicefrac\beta2, \nicefrac\rho3)$ using $M_1$ samples
    \STATE Draw a dataset $S\sim \mcal D^{M_2}$ of size $M_2$.
    \FOR {each leaf $t \in T$}
        \IF {$\hat{\mcal D}(t\mid T) \geq \nicefrac{\alpha}{4\cdot 2^\ell}$}
            \STATE Let $S_t$ be the subset of samples in $S$ that reach $t$.
            \STATE Create a set $S_t'$ consisting of points in $S_t$ but where all coordinates queried on the root-to-leaf path for $t$ are re-randomized independently, making the marginal distribution uniform.
            \STATE For $\rho'\in (0, 1)$,
            choose its parameters so that $\mcal A = \mcal A(\nicefrac\alpha6, \rho', \nicefrac16)$ is $\rho'$-replicable and $(\nicefrac\alpha6, \nicefrac16+\nicefrac13, c')$-robustly learns $\mscr C$.
            \STATE Get $h_t$ by calling $\rBoost_{\mathcal A}(S_t', \nicefrac\alpha6, \nicefrac\rho{3\cdot 2^{\ell}}, \nicefrac\beta{2\cdot 2^{\ell}})$ (cf. \Cref{prop:replicable boosting}).
        \ELSE
            \STATE Get $h_t$ that outputs a random guess according to shared randomness.
        \ENDIF    
    \ENDFOR
    \STATE Return the hypothesis $h$ such that given input $x$, 
    finds the leaf $t \in T$ that $x$ follows and outputs $h_t(x)$.
    \end{algorithmic}
\label{algorithms:lift}
\end{algorithm}

{Now,
our learning algorithm is designed for an exact tree distribution.
However,
we can only estimate the tree representation of the actual distribution,
i.e., the conditional distribution at a leaf subcube is uniform in the tree representation
but not necessarily uniform in the input distribution.
To show that our learning algorithm can accommodate slight errors in the tree distribution estimation,
we introduce the following definition.}
\begin{defn}
[Robust Learning]
For any concept class $\mathscr C$ and algorithm $\mathcal A$, we say that $\mathcal A$ $(\alpha, \beta, c)$-robustly learns $\mathscr C$ using $m$ samples under the uniform distribution if for any $\eta > 0$ and class
\[
\mathscr D_\eta = \{ \text{distribution $\mathcal{D}$ over $\Set{0, 1}^d$} ~:~ \TV(\mathcal{U}, \mathcal{D}) \leq \eta  \}\,,
\]
it holds that
$\mcal A(\alpha + c \eta, \beta)$-learns $\mathscr C$ using $m$ samples
with respect to the distributions in
$\mathscr D_\eta.$ 
\end{defn}

\begin{prop}[Proposition 7.2 from \citep{blanc2022properly}]\label{prop:uniform robust learning}
    For any concept class $\mscr C$ and algorithm $\mcal A$,
    if $\mcal A$ $(\alpha, \beta)$-PAC learns $\mscr C$ using $m$ samples under the uniform distribution,
    then the exact same algorithm $\mcal A$ also $(\alpha, \beta+\nicefrac13, 3m)$-robustly learns $\mscr C$ using $m$ samples.
\end{prop}

We are now ready to state the main result of this section,
from which the proof of \Cref{thm:lift} closely follows.
\begin{thm}
\label{thm:robust-lift}
Consider a concept class $\mathscr{C}$ of functions $f: \{0,1\}^d \to \{0,1\}$ closed under restrictions.
Fix $\alpha, \beta, c, \rho > 0$
and $m,\ell \in \mathbb N$.
Suppose we are provided black-box access to an algorithm $\mathcal A$ that 
\begin{enumerate}[(i)]
    \item is $\rho'$-replicable with respect to the uniform distribution for any $\rho'\in (0, 1)$,
    \item $(\alpha', \beta'+\nicefrac13, c)$-robustly learns $\mathscr{C}$ for any $\alpha', \beta'\in (0, 1)$, and
    \item consumes $m(\alpha', \rho', \beta') = \poly(d, \nicefrac1{\alpha'}, \nicefrac1{\rho'}, \log(\nicefrac1{\beta'}))$ samples 
    and computation time under the uniform distribution.
\end{enumerate}
Let $M_1 = \poly(2^\ell, \nicefrac1\alpha, \nicefrac1\rho, \log(\nicefrac1\beta))$,
$M_2 = m(\nicefrac\alpha6, \nicefrac\rho{2^\ell}, \nicefrac16)\cdot \poly(d, 2^\ell,\nicefrac1\alpha, \nicefrac1\rho, \log(\nicefrac1\beta))$,
and
\begin{align*}
    M &:= M_1 + M_2
    \leq \poly(d, 2^\ell,\nicefrac1\alpha, \nicefrac1\rho, \log(\nicefrac1\beta)).
\end{align*}
For any function $f^\star \in \mathscr C$, distribution $\mathcal D$ over $\{0,1\}^d$, depth-$\ell$ decision tree $T$ computing the pmf of a distribution $\mathcal D_T$ where
$
\TV(\mathcal{D}, \mathcal{D}_T) \leq \nicefrac\alpha{6c}
$,
the following holds.

The algorithm $\rLift(T, \mathcal{A}, \mcal D, \alpha, \beta, \rho)$ (cf. \Cref{algorithms:lift})
is $\rho$-replicable,
has $O(M)$ sample complexity,
terminates in $\poly(M)$ time,
and its output is $\alpha$-close
to $f^\star$ with respect to $\mathcal D$ with probability at least $1-\beta$.
\end{thm}

Before we prove \Cref{thm:robust-lift},
we state two useful lemmas.

\begin{lem}

\label{lem:reaches size}
    \sloppy
    Fix $\alpha, \rho, \rho', \beta\in (0, 1)$,
    $\Delta \geq 0$,
    and $\ell\in \N$.
    Let $m_\rBoost = \poly(d, 2^\ell, \nicefrac1\alpha, \nicefrac1\rho, \log(\nicefrac1\beta))$
    be the number of samples that $\rBoost$ (cf. \Cref{prop:replicable boosting}) requires to boost a $\rho'$-replicable $(\nicefrac\alpha6 + \Delta, \nicefrac12)$-correct learner (that uses $\poly(d, \nicefrac1{\alpha}, \nicefrac1{\rho'})$ samples and running time)
    to a $\nicefrac\rho{3\cdot 2^{\ell}}$-replicable learner with accuracy $\nicefrac\alpha6 + \Delta  + \nicefrac\alpha6 = \nicefrac\alpha3 + \Delta$ and confidence $\nicefrac\beta{2\cdot 2^{\ell}}$.
    Condition on the success of \rFiniteDistrEst in \Cref{algorithms:lift}.
    Then with probability at least $1-\nicefrac\beta2$,
    for every $t \in T$ with $\hat{\mcal D}(t\mid T)\geq \nicefrac{\alpha}{4\cdot 2^\ell}$,
    we observe at least $m_\rBoost$ samples reaching $t$ from $S$.
\end{lem}

\begin{pf}
    Condition on the success of the call to \rFiniteDistrEst.
    Then for every $t \in T$ with $\hat{\mcal D}(t\mid T)\geq \nicefrac{\alpha}{4\cdot 2^\ell}$,
    we have $\mcal D(t\mid T) \geq \nicefrac{\alpha}{6\cdot 2^\ell}$.
    The probability of not observing such a $t$ in a dataset of size $n$
    is at most $(1-\nicefrac{\alpha}{6\cdot 2^\ell})^n \leq \exp(-\nicefrac{n\alpha}{6\cdot 2^\ell})$.
    By a union bound over the at most $\nicefrac{6\cdot 2^\ell}\alpha$ such $t$,
    the probability of failing to collect each such $t$ is at most
    \[
        \frac{6\cdot 2^\ell}\alpha\exp(-\nicefrac{n\alpha}{6\cdot 2^\ell}).
    \]
    To drive this below the constant $\nicefrac12$,
    it suffices to take $n = O(\nicefrac{\ell\cdot 2^\ell}\alpha \log\nicefrac1\alpha)$.
    By \Cref{prop:multiple coupon collector},
    it suffices to take a sample of size
    \[
        O(nm_\rBoost \log\nicefrac1\beta)
        = O\left( \frac{m_\rBoost \ell\cdot 2^\ell}\alpha \left( \log\frac1\alpha \right) \left( \log\frac1{\beta} \right) \right)
        \leq M_2
    \]
    to collect $m$ copies of each such $t$
    with probability at least $1-\nicefrac\beta2$.
\end{pf}

\begin{lem}[Lemma B.4 from \citep{blanc2022popular}, Fact 7.4 from \citep{blanc2023lifting}]\label{lem:reaches TV}
    For any distribution $\mcal D$ and decision tree $T$ computing the pmf of another distribution $\mcal D_T$,
    \[
        \sum_{t\in T} \Pr_{x\sim \mcal D}\left[ \text{$x$ reaches $t$} \right]\cdot \TV(\mcal D_t, (\mcal D_T)_t)
        \leq 2\TV(\mcal D, \mcal D_T)\,,
    \]
    where $(\mcal D_T)_t$ (resp. $\mcal D_t)$ is the conditional of $\mcal D_T$ (resp. $\mcal D)$ on the subcube induced by the leaf $t$.
\end{lem}

Note that in the statement of the lemma above,
$\Pr_{x\sim \mcal D}\left[ \text{$x$ reaches $t$} \right] = \mcal D(t\mid T)$
and $(\mcal D_T)_t$ is the uniform distribution on the subcube represented by the leaf $t$.
We are finally ready to prove \Cref{thm:robust-lift}.
\begin{pf}
[\Cref{thm:robust-lift}]
We first note that the sample complexity of the call to the routine \rFiniteDistrEst is $M_1 = \poly(2^\ell, \nicefrac1\alpha, \nicefrac1\rho, \log(\nicefrac1\beta))$ (cf. \Cref{cor:replicable mass estimation}).
Throughout this proof,
we condition on the $1-\nicefrac\beta2$ probability of \rFiniteDistrEst succeeding.

Let $h$ be the output of $\rLift(T, \mathcal{A}, S)$.
We now argue separately about the accuracy and replicability of the algorithm.

\paragraph{Accuracy.} The accuracy of the learner $h = \rLift(T, \mathcal{A}, S)$
is analyzed as follows.
For any leaf $t$ of the tree $T$,
let $h_t$ be the associated predictor.
We have that
\[
\Pr_{x \sim \mathcal D}
[h(x) \neq f^\star(x)]
=
\sum_{t \in T}
\Pr_{x \sim \mathcal D}
[x ~\text{reaches}~ t]
\Pr_{x \sim \mathcal{D}_t}
[h_t(x) \neq f^\star(x)]\,.
\]
Each hypothesis $h_t$ is obtained either by running 
$\rBoost_{\mathcal{A}}$ (cf. \Cref{prop:replicable boosting}) on the sample $S_t'$ given that $\hat{\mcal D}(t\mid T) \geq \nicefrac{\alpha}{4\cdot 2^\ell}$
or corresponds to a random guess otherwise.

On the other hand,
from the choice of estimation error to the call for \rFiniteDistrEst,
each leaf $t$ on which we output the random guess hypothesis
must satisfy $\Pr_{x\sim \mcal D}[\text{$x$ reaches $t$}] < \nicefrac\alpha{3\cdot 2^\ell}$.
The overall error contribution of such leaves is thus at most $\nicefrac\alpha3$.
It follows that total error is at most
\begin{align*}
    \Pr_{x \sim \mathcal D}
    [h(x) \neq f^\star(x)]
    &\leq
    \sum_{t \in T: \hat{\mcal D}(t\mid T) \geq \nicefrac{\alpha}{4\cdot 2^\ell}}
    \Pr_{x \sim \mathcal D}
    [x ~\text{reaches}~ t]
    \Pr_{x \sim \mathcal{D}_t}
    \left[ h_t(x) \neq f^\star(x) \right] + \frac\alpha3 \,.
\end{align*}

Fix a $t\in T$ on which we run $\rBoost_{\mcal A}$.
Recall that $\mcal D_t$ and $(\mcal D_T)_t$ denotes the conditional distribution on the leaf subcube 
{over the coordinates not fixed by $\pi = \pi(t)$ where $\pi$ is the restriction corresponding to $t$}.
Here the underlying distributions are the input distribution $\mcal D$ and tree distribution $\mcal D_T$.
Let $\mcal D_t', (\mcal D_T)_t'$ denote the distributions over $\Set{0, 1}^d$
obtained from $\mcal D_t, (\mcal D_T)_t$ by re-randomizing the coordinates from $\pi$.
Then $(\mcal D_T)_t'$ is precisely the uniform distribution $\mcal U$ over $\Set{0, 1}^d$
and $\mcal D_t'$ satisfies
\begin{align*}
    &\TV(\mcal D_t', \mcal U) \\
    &= \TV(\mcal D_t', (\mcal D_T)_t') \\
    &:= \frac12 \sum_{x\in \Set{0, 1}^d} \abs{\mcal D_t'(x) - (\mcal D_T)_t'(x)} \\
    &= \frac12 \sum_{x\in t} 2^{\card{\pi(t)}} \abs{\mcal D_t'(x) - (\mcal D_T)_t'(x)} \\
    &= \frac12 \sum_{x\in t} \abs{\mcal D_t(x) - (\mcal D_T)_t(x)} 
    &&\mcal D_t'(x) = 2^{-\card{\pi(t)}}\mcal D_t(x),
    (\mcal D_T)_t'(x) = 2^{-\card{\pi(t)}}(\mcal D_T)_t(x) \\
    &= \TV(\mcal D_t, (\mcal D_T)_t).
\end{align*}
Based on the re-sampling step of the algorithm, 
we know that
any point in $S_t'$ is an i.i.d. sample from $\mcal D_t'$
and labeled by the function $f^\star_t$, 
which lies in $\mathscr C$ thanks to closedness under restrictions.

By $(\nicefrac\alpha6, \nicefrac16+\nicefrac13, c)$-robust learnability,
$\mcal A$ in fact $(\nicefrac\alpha6 + c\TV(\mcal D_t', \mcal U), \nicefrac12)$-PAC learns $\mscr C$ under the distribution $\mcal D_t'$.
We now apply \Cref{lem:reaches size} with $\Delta = c\TV(\mcal D_t', \mcal U)$:
the leaves on which we run $\rBoost_{\mcal A}$ have at least $m_\rBoost$ samples 
where $m_\rBoost = \poly(d, 2^\ell, \nicefrac1\alpha, \nicefrac1\rho, \log(\nicefrac1\beta))$
is the number of samples that $\rBoost$ requires to boost a $\rho'$-replicable $(\nicefrac\alpha6 + c\TV(\mcal D_t', \mcal U), \nicefrac12)$-correct learner 
(that uses $\poly(d, \nicefrac1{\alpha}, \nicefrac1{\rho'})$ samples and running time)
to a $\nicefrac\rho{3\cdot 2^{\ell}}$-replicable learner with accuracy $\nicefrac\alpha6 + c\TV(\mcal D_t', \mcal U) + \nicefrac\alpha6 = \nicefrac\alpha3 + c\TV(\mcal D_t', \mcal U)$ and confidence $\nicefrac\beta{2\cdot 2^{\ell}}$.

Then by \Cref{prop:replicable boosting},
with probability at least $1 - \nicefrac\beta{2\cdot 2^{\ell}}$ over the data $S_t'$, 
the hypothesis $h_t$ output by $\rBoost_{\mcal A}$ satisfies 
\[
    \Pr_{x \sim \mathcal{D}_t}
    \left[ h_t(x) \neq f^\star(x) \right]
    \leq \nicefrac\alpha3 + c~ \TV(\mathcal{D}_t', \mathcal{U})
    = \nicefrac\alpha3 + c~ \TV(\mathcal{D}_t, (\mcal D_T)_t) \,.
\]

All in all,
Combined with the expression of total error above,
this means that with probability at least $1-\beta$,
\[
\Pr_{x \sim \mathcal{D}}
[h(x) \neq f^\star(x)]
\leq 
\nicefrac{2\alpha}3 +
c \sum_{t \in T}
\Pr_{x \sim \mathcal{D}}[x~\text{reaches}~t]~
\TV(\mathcal{D}_t, (\mcal D_T)_t)\,.
\]
We finish by applying \Cref{lem:reaches TV},
which ensures that the sum over leaves above is upper bounded by $2\TV(\mathcal{D}, \mathcal{D}_T) \leq \nicefrac\alpha{3c}$ and so the total misclassification error is at most $\alpha$ with probability at least $1-\beta$.

\paragraph{Replicability.} 
We assume we are given the same decision tree in both executions. 
Next,
we note that \rFiniteDistrEst is $\nicefrac\rho3$-replicable
and succeeds in two executions with probability at least $1-2\cdot\nicefrac\beta2\geq 1-\nicefrac\rho3$.
Conditional on the events above,
it suffices to show that for any leaf of the tree, the algorithm $\rBoost_{\mathcal A}$ replicably outputs a hypothesis. 
Since there are at most $2^\ell$ leaves and each call of $\rBoost_{\mathcal{A}}$ has replicability parameter $\nicefrac\rho{3\cdot 2^{\ell}}$, the entire procedure is $\rho$-replicable.
\end{pf}

Before moving on to the proof of \Cref{thm:lift},
we state one final lemma.
\begin{lem}\label{lem:replicable robust learning}
    For any concept class $\mathscr C$ and algorithm $\mathcal A$, if
    $\mathcal A$ $\rho$-replicably $(\alpha, \beta)$-learns $\mathscr C$ using $m$ samples under the uniform distribution,
    then
    the exact same algorithm $\mathcal A$ also
    $\rho$-replicably
    $(\alpha, \beta+\nicefrac13, 3m)$-robustly learns $\mathscr C$ using $m$ samples 
    under the uniform distribution.
\end{lem}

\begin{pf}[\Cref{lem:replicable robust learning}]
    The proof follows directly from the analysis of Proposition 7.2 in \citet{blanc2023lifting} (cf. \Cref{prop:uniform robust learning}). 
    The replicability of the algorithm is trivially preserved since the algorithm does not change.
\end{pf}

We restate the statement of \Cref{thm:lift} below for convenience
before its formal proof.
\rLiftUniformLearners*
\begin{pf}
[\Cref{thm:lift}]
    Using \Cref{thm:robust-lift}, 
    we can prove \Cref{thm:lift} as follows. First, we use \Cref{thm:learn-DT} to replicably learn the input distribution to total variation distance $\nicefrac\alpha{6\cdot 3m}$ with a decision tree. 
    Next,
    \Cref{lem:replicable robust learning} asserts that ${\mcal A}$ is $\rho$-replicable and $(\alpha, \beta+\nicefrac13, 3m)$-robustly learns $\mathscr C$ under the uniform distribution.
    The result follows by applying \Cref{thm:robust-lift}.
\end{pf}

\section{Efficient Replicability and Private Learning}
\label{apx:private to replicable}
\subsection{Replicably Learning Finite Classes}\label{apx:omitted-dp-transformation}
We first state a useful result which we later leverage as a subroutine appears in \cite{bun2023stability}.
\begin{thm}[Replicable Learner for Finite Classes; Theorem 5.13 in \citep{bun2023stability}]
\label{thm:agnostic-replicable-learner-finite-class}
    Let $\alpha, \rho, \beta \in (0,1)^3$. Then, any finite
    concept class $\mscr C$ is replicably learnable in the agnostic
    setting with
    \[
        m(\alpha,\rho, \beta) = O\left(\frac{\log^2|\mscr C| + \log\frac{1}{\rho\beta}}{\alpha^2\rho^2} \log^3\frac{1}{\rho}\right) \,,
    \]
    many samples. Moreover, the running time of the algorithm is polynomial in the number of samples and the cardinality of $\mscr C.$
\end{thm}

\subsection{A Transformation from Pure DP to Replicability}\label{sec:pure-dp-to-replicability}
Our approach 
borrows some high level ideas from \citet{gonen2019private} who showed a transformation
from a pure DP PAC learner to an online learner. In a nutshell, our algorithm
works as follows:
\begin{enumerate}[1)]
    \item First, we create a ``dummy'' input dataset $\bar S$ which has some constant size\footnote{As in \citet{gonen2019private}, the size is constant w.r.t. the accuracy, confidence parameters, but it might not be constant w.r.t. some parameter the describes the complexity of the concept class. }
    and we run the algorithm for $\widetilde{\Theta}\left( \frac{\log^3(\nicefrac1\beta) }{\rho^2}\right)$ times 
    on $\bar S$, resampling its internal randomness every time.

    \item With probability $1-\beta$, one of the hypotheses will have error at most $\nicefrac38$ \citep{beimel2013characterizing, gonen2019private}.

    \item We run the replicable agnostic learner for finite classes
    from \cite{bun2023stability}.

    \item We boost the weak replicable learner using \citet{impagliazzo2022reproducibility, kalavasis2023statistical}.
\end{enumerate}

We begin by presenting an adaptation of a result that appeared in \citet{beimel2013characterizing, gonen2019private}.
Essentially, it states that
when a class is learnable by a pure DP learner, one can fix an  
arbitrary input $\bar S$ that has constant size, execute the learner on this dataset
a constant number of times by resampling its \emph{internal} randomness, and
then with some constant probability, e.g., $\nicefrac{15}{16}$, the output will contain
one hypothesis whose error is bounded away from $\nicefrac12$ by some constant, e.g., $\nicefrac14.$
This result might seem counter-intuitive at first glance, 
but learnability by a pure DP learner is a very strong property for a class
and the result crucially depends on this property. 
Below, we present an adaptation of this result which states that the
weak learner can be $\rho$-replicable and has a probability of success of $1-\delta$. 
An important element of our derivation is a result from \citet{bun2023stability} about the sample and computational complexity of agnostically
learning a concept class using a replicable 
algorithm (cf. \Cref{thm:agnostic-replicable-learner-finite-class}).

We are now ready to state a key lemma for our transformation.
\begin{lem}[From Pure DP Learner to Replicable Weak Learner]\label{lem:pure-dp-weak-replicable}
    Let $\mcal X$ be some input domain, $ \mcal Y = \{0,1\}$,
    and $\mcal D_{XY}$ be a distribution
    on $\mcal X \times \mcal Y$ that is realizable with respect to some 
    concept class $\mscr C$. Let $\mcal A$ be a pure DP learner such that for
    any $\alpha, \varepsilon, \beta \in (0,1)^3$,
    $\mcal A$ needs $m(\alpha, \varepsilon, \beta, \mscr C)$ i.i.d. samples from $\mcal D_{XY}$ 
    to output with probability at least $1-\beta$ a hypothesis with error at most $\alpha$ in an $\varepsilon$-DP way.
    Let us set $m_0 = m(\nicefrac14, \nicefrac1{10}, \nicefrac12, \mscr C).$
    
    Then, for any $\rho,\beta' \in (0,1)^2$
    there is a $\rho$-replicable learner $\mcal A'$
    that outputs a hypothesis with error at most $\nicefrac38$ with probability at 
    least $1-\beta'$ and
    requires $\widetilde{O}\left( \poly(m_0) \frac{\log^3(\nicefrac1{\beta'}) }{\rho^2}\right)
    $ 
    i.i.d.
    samples from $\mcal D_{XY}$ and $O(\exp(m_0)\log(\nicefrac1{\beta'}))
    $ oracle calls to $\mcal A$ (and hence runtime).
\end{lem}

\begin{pf}[\Cref{lem:pure-dp-weak-replicable}]
    We argue about the correctness and replicability of the algorithm separately.
    
    \paragraph{Correctness.}
    Since $\mcal D_{XY}$ is realizable,
    there exists some $c^\star \in \mscr C$ such that
    $\loss_{\mcal D_{XY}}(c^\star) = 0.$ Let 
    \[
        \mathrm{range}(\mcal A) = \{c:\mcal X \rightarrow \{0,1\}: \exists S \in \cup_{n \in \mathbb{N}} (\mcal X \times \{0,1\})^n, \exists r \in \cup_{n \in \mathbb{N}} \{0,1\}^n, \text{ s.t. } \mcal A(S;r) = c \} \,.
    \]
    Let also $m_0 = m(\nicefrac14, \nicefrac1{10}, \nicefrac12, \mscr C)$ be the number of samples $\mcal A$ needs to achieve $\alpha = \nicefrac14, \varepsilon = \nicefrac1{10}, \beta = \nicefrac12$ for the class $\mscr C$. 
    Notice that $m_0$ is a constant with respect to $\alpha,\beta,\rho,\epsilon$.
    Let also
    \[
        \mscr C(\mcal D_{XY}) = \{c \in \mathrm{range}(\mcal A): \loss_{\mcal D_{XY}}(c) \leq 1/4\} \,.
    \]
    By definition, with probability at least $\nicefrac12$ over the random draw of an i.i.d. sample
    of size $m_0$ from $\mcal D_{XY}$ and the internal randomness of $\mcal A$, the output
    of the algorithm belongs to $\mscr C(\mcal D_{XY}).$ 
    Thus, there exists a sample
    $S_0 \in (\mcal X \times \{0,1\})^{m_0}$ such that 
    \[
        \Pr_{r\sim \mcal R}[\mcal A(S_0;r) \in \mscr C(\mcal D_{XY})] \geq 1/2 \,.
    \]
    Although it is unclear how to identify such a $S_0$,
    notice that any sample $S \in (\mcal X \times \{0,1\})^{m_0}$ 
    is an $m_0$-neighbor of $S_0$.
    Since $\mcal A$ is pure DP,
    it holds that
    \[
        \Pr_{r\sim \mcal R}[\mcal A(S;r) \in \mscr C(\mcal D_{XY})] \geq 1/2\cdot e^{-0.1\cdot m_0} \,.
    \]
    In particular,
    this holds for the dummy dataset $\bar S$ we fix before observing any data.
    Then if we sample $N = 2\cdot e^{0.1m_0} \cdot \log(\nicefrac3{\beta'}) = \Theta(\exp(m_0) \cdot \log(\nicefrac1{\beta'}))$ i.i.d.
    random strings $r_1,\ldots,r_N$ from $\mcal R$ i.i.d. the probability that none of the hypotheses
    $\{ \mcal A(\bar S; r_i) \}_{i \in [N]}$ is in $\mscr C(\mcal D_{XY})$ is at most
    \[
        (1 - \nicefrac12\exp(-0.1m_0))^{2\exp(0.1m_0) \cdot \log(\nicefrac{1}{\beta'})} \leq \frac{\beta'}3 \,.
    \]
    We denote the event that one of the outputs is in $\mscr C(\mcal D_{XY})$ by $\mcal E$
    and we condition on it for the rest of the correctness proof. The next step is
    to replicably learn the finite concept class $\bar{\mscr C} = \{\mcal A(\bar S; r_1),\ldots,\mcal A(\bar S; r_N)\}$ we have
    constructed using \Cref{thm:agnostic-replicable-learner-finite-class} with accuracy $\nicefrac18$, confidence $\nicefrac{\beta'}3$,
    and replicability $\rho$. Thus, we can see that $O\left(\frac{\log^2N + \log\frac{1}{\beta\rho}}{\rho^2} \log^3\frac{1}{\rho}\right) = \widetilde O\left(\frac{m_0^2\log(1/\beta')}{\rho^2}\right)$
    samples suffice for this task. Let $\mcal E'$ be the event that this estimation is correct, which happens with probability at least $\nicefrac{\beta'}3.$ Conditioned on that event, 
    outputting the hypothesis generated by the learner gives a solution with error 
    at most $\nicefrac14 + \nicefrac18 = \nicefrac38$. Notice that the good events happen with probability at least $1-\nicefrac{2\beta'}3 \geq 1-\beta,$ so the correctness condition is satisfied.
    
    \paragraph{Replicability.} Notice that the first step of the algorithm 
    where we fix a dummy dataset $\bar S$ of size $m_0$ and run the algorithm on i.i.d.
    strings of its internal randomness is trivially replicable, since the dataset and
    the randomness are the same across the two executions. Then, by the guarantees of 
    \Cref{thm:agnostic-replicable-learner-finite-class},
    we know that the output of the algorithm will be the same
    across two executions
    with probability at least $1-\rho$. Thus, overall we see that with probability at least $1-\rho$,
    the output of our algorithm is the same across the two executions.
\end{pf}

We emphasize that the number of oracle calls to the DP algorithm is constant with respect to the confidence and the 
correctness parameter, but could, potentially,
be exponential with respect to some parameter that depends
on the representation of the underlying concept
class $\mcal X.$

Equipped with the weak learner from \Cref{lem:pure-dp-weak-replicable} we can
boost its error parameter using the boosting algorithm
that appears in \citet{impagliazzo2022reproducibility}. For completeness, we state the
result below.

\begin{thm}[Replicable Boosting Algorithm \citep{impagliazzo2022reproducibility}]
    Let $\mcal D_{XY}$ be the joint distribution over labeled examples as in \Cref{lem:pure-dp-weak-replicable}.
    Fix $\alpha, \rho, \beta' > 0$.
    Let $\mcal A$ be a $\rho$-replicable $(\gamma, \beta)$-weak learner with sample complexity $m(\gamma, \rho, \beta)$, 
    i.e., its error is at most $\nicefrac{1}{2}-\gamma$,
    with probability at least $1-\beta$.
    Then, there exists an efficient
    $\rho$-replicable boosting algorithm $\mcal A'$ such that
    with probability at least $1-\beta'$ it outputs a hypothesis $h$ with $\loss_{\mcal D_{XY}}(h) \leq \alpha.$ The algorithm runs for $T = \widetilde{O}(\log(\nicefrac1\beta)/(\alpha \rho^2)) $ rounds and requires
    $T$ oracle calls to the weak learner. Moreover, it requires
    \[
        \widetilde{O}\left( \left( \frac{m(\gamma, \nicefrac\rho{6T}, \beta)}{\alpha^2\gamma^2} + \frac{1}{\rho^2\alpha^3\gamma^2} \right)\log\frac1{\beta'} \right)
    \]
    many samples, 
    where $m(\gamma, \nicefrac\rho{6T}, \beta)$ 
    is the sample complexity needed for the weak learner to be $\nicefrac\rho{6T}$-replicable
    and $(\nicefrac12-\gamma, \beta)$-accurate.
\end{thm}

Combining the discussion above, 
we see that for any $\alpha, \rho, \beta \in (0,1)^3$
we can transform a pure DP learner to a $\rho$-replicable $(\alpha,\beta)$-correct (strong)
learner
{where the transformation is efficient with respect to the accuracy $\alpha$,
confidence $\beta$,
and replicability $\rho$.}
This proves \Cref{thm:efficient-pure-dp-to-replicable},
which we restate below for convenience.
\dpToReplicable*

We reiterate that even though our transformation is 
efficient with respect to the input parameters $\alpha,\beta,\rho,$
it might not be efficient with respect to some parameter that depends
on the 
representation of the concept class. This is because the size
of the ``dummy'' dataset $m_0$ could depend
on the size of that representation and our transformation
requires $\poly(m_0)$ samples but $\exp(m_0)$ running time. 
This is also the case for transformation from a DP learner
to an online learner \citep{gonen2019private}.

\end{document}